\documentclass[10pt,twocolumn,letterpaper]{article}

\usepackage[pagenumbers]{cvpr} % To force page numbers, e.g. for an arXiv version

\usepackage{graphicx}
\usepackage{amsmath}
\usepackage{amssymb}
\usepackage{booktabs}
\usepackage{enumitem}
\usepackage{thmtools}
\usepackage{thm-restate}

\newtheorem{theorem}{Theorem}[section]
\newtheorem{lemma}[theorem]{Lemma}

\usepackage[pagebackref,breaklinks,colorlinks]{hyperref}

\usepackage[capitalize]{cleveref}
\crefname{section}{Sec.}{Secs.}
\Crefname{section}{Section}{Sections}
\Crefname{table}{Table}{Tables}
\crefname{table}{Tab.}{Tabs.}

\usepackage{xcolor}

\def\cvprPaperID{5972} % *** Enter the CVPR Paper ID here
\def\confName{CVPR}
\def\confYear{2022}

\begin{document}

\title{Can Neural Nets Learn the Same Model Twice? Investigating Reproducibility and Double Descent from the Decision Boundary Perspective}

% \author{Gowthami Somepalli\\
% University of Maryland, College Park\\
% {\tt\small gowthami@umd.edu}
% \and
\author{Gowthami Somepalli \textsuperscript{\rm 1}, 
Liam Fowl  \textsuperscript{\rm 1}, 
Arpit Bansal \textsuperscript{\rm 1}, 
Ping Yeh-Chiang \textsuperscript{\rm 1},
Yehuda Dar \textsuperscript{\rm 2},\\
Richard Baraniuk \textsuperscript{\rm 2},
Micah Goldblum \textsuperscript{\rm 3},
Tom Goldstein \textsuperscript{\rm 1}
\\
\and 
\textsuperscript{\rm 1} University of Maryland, College Park\\
{\tt\small \{gowthami, pchiang, tomg\}@cs.umd.edu}\\
{\tt\small lfowl@math.umd.edu}, {\tt\small bansal01@umd.edu}
\and
\textsuperscript{\rm 2} Rice University\\
{\tt\small \{ydar, richb\}@rice.edu}
\and
\textsuperscript{\rm 3} New York University\\
{\tt\small goldblum@nyu.edu}
}

\maketitle

%%%%%%%%% ABSTRACT
\begin{abstract}
   We discuss methods for visualizing neural network decision boundaries and decision regions. We use these visualizations to investigate issues related to reproducibility and generalization in neural network training.  We observe that changes in model architecture (and its associate inductive bias) cause visible changes in decision boundaries, while multiple runs with the same architecture yield results with strong similarities, especially in the case of wide architectures.  We also use decision boundary methods to visualize double descent phenomena. We see that decision boundary reproducibility depends strongly on model width. Near the threshold of interpolation, neural network decision boundaries become fragmented into many small decision regions, and these regions are non-reproducible.  Meanwhile, very narrows and very wide networks have high levels of reproducibility in their decision boundaries with relatively few decision regions. We discuss how our observations relate to the theory of double descent phenomena in convex models. Code is available at \url{https://github.com/somepago/dbViz} .
\end{abstract}

%%%%%%%%% BODY TEXT
\section{Introduction}
\label{sec:intro}
The superiority of neural networks over classical linear classifiers stems from their ability to slice image space into complex class regions. While neural network training is certainly not well understood, existing theories of neural network training mostly focus on understanding the geometry of loss landscapes~\cite{chaudhari2019entropy,dinh2017sharp,li2017visualizing}. Meanwhile, considerably less is known about the geometry of class boundaries. The geometry of these regions depends strongly on the inductive bias of neural network models, which we do not currently have tools to rigorously analyze.  To make things worse, the inductive bias of neural networks is impacted by the choice of architecture, which further complicates theoretical analysis.

\begin{figure}[t]
\begin{center}
\includegraphics[width=\linewidth]{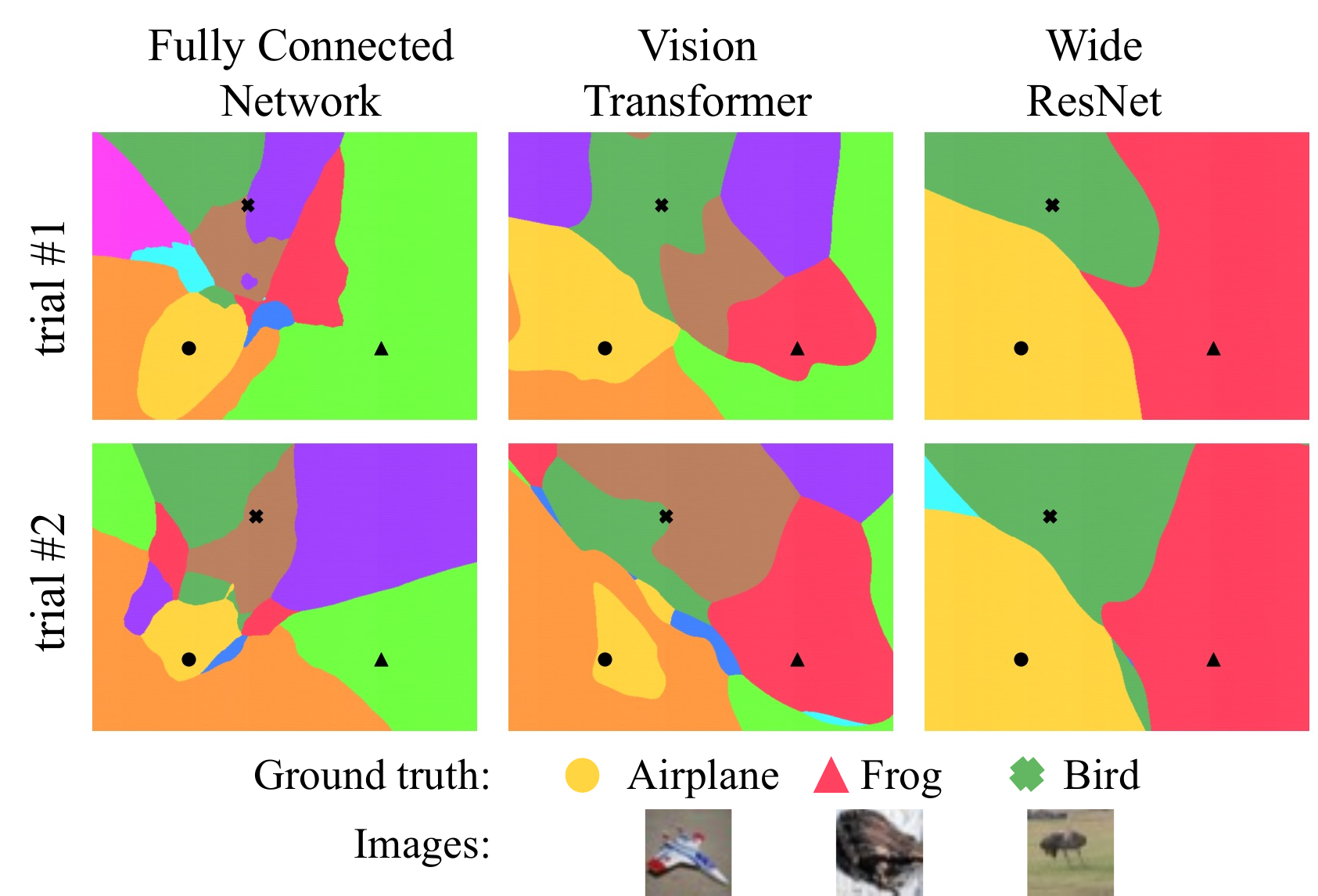}
\end{center}
\vspace{-5mm}
\caption{
The class boundaries of three architectures, plotted on the plane spanning three randomly selected images.  Each model is trained twice with random seeds.  Decision boundaries are reproducible across runs, and there are consistent differences between the class regions created by different architectures.}
\label{fig:teaser}
\end{figure}

In this study, we use empirical tools to study the geometry of class regions, and how neural architecture impacts inductive bias.  We do this using visualizations and quantitative metrics calculated using realistic models. We start by presenting simple methods for decision boundary visualization. Using visualization as a tool, we do a deep dive on three main issues:
\begin{itemize}[noitemsep,nolistsep]
    \item  Do neural networks produce decision boundaries that are consistent across random initializations?  Put simply, can a neural network learn the same model twice? We see empirically that the decision boundaries of a network have strong similarities across runs, and we confirm this using quantitative measurements.
    \item Do different neural architectures have measurable differences in inductive bias?  Indeed, we find clear visible differences between the class regions of different model architectures (e.g., ResNet-18 vs ViT). 
    \item We use decision boundary visualizations to investigate the ``double descent'' phenomenon.  We see that decision boundaries become highly unstable and fragmented when model capacity is near the interpolation threshold, and we explore how  double descent in neural networks relates to known theory for linear models. 
\end{itemize}

\section{Plotting decision boundaries} 
Most prior work on decision boundary visualization is for the purpose of seeing the narrow margins in adversarial directions \cite{karimi2019characterizing,he2018decision}. Fawzi et al.~\cite{fawzi2018empirical} visualize the topological connectivity of classification regions.
To facilitate our studies, we seek a general-purpose visualization method that is simple, controllable, and captures import parts of decision space that lie near the data manifold. 

\subsection{On-manifold vs off-manifold behavior}
\begin{figure}[t]
    \centering
    \vspace{-1mm}
    \includegraphics[width=0.9\linewidth]{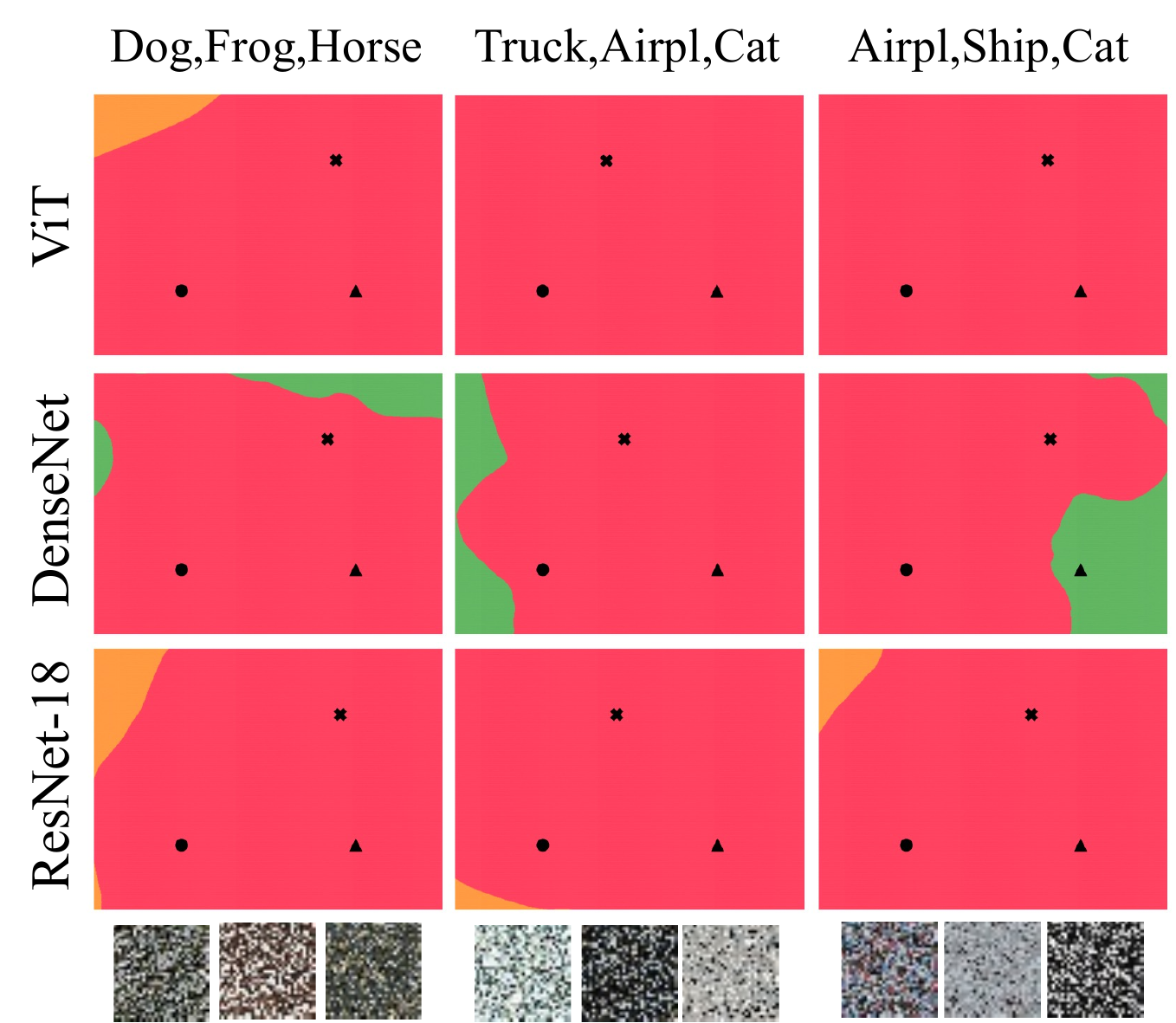}
    \vspace{-2mm}
    \caption{Off-manifold decision boundaries near ``random'' images created by shuffling pixels in CIFAR-10 images. Each column's title shows the labels of the unshuffled base images. Below each column we show the shuffled image triplet. Color-class mapping is as follow Red:Frog, Green:Bird, Orange:Automobile.}
    \vspace{-4mm}
    \label{fig:dd_random_shuffle}
\end{figure}

When plotting decision boundaries, it is important to choose a method that captures the behavior of models near the data manifold.  To understand why, consider the plots of decision boundaries through planes spanning randomly chosen points in input space as shown in Figure~\ref{fig:dd_random_shuffle}.  We see that decision regions are extremely smooth and uniform with few interesting features. The training process, which structures decision boundaries near the data manifold (e.g. Fig.~\ref{fig:teaser}), fails to produce strong structural effects far from the manifold (e.g. Fig.~\ref{fig:dd_random_shuffle}).   

The uniform off-manifold behavior is not particular to our training method or architecture but is rather an inevitable consequence of the concentration of measures phenomenon \cite{shafahi2018adversarial,ledoux2001concentration}.  In fact, we can show that any neural network that varies smoothly as a function of its input will assume nearly constant outputs over most of input space.  The proof of the following result is in Appendix~\ref{appendix_sec:lemma_proof}.
%\begin{lemma}
\begin{restatable}{lem}{mainlemma}
\vspace{-1mm}Let $f:[0,1]^n \to [0,1]$ be a neural network satisfying $|f(x) - f(y)| \le \frac{L}{\sqrt{n}}\|x-y\|.$ Let $\bar{f}$ denote the median value of $f$ on the unit hypercube. Then, for an image $x\in [0,1]^n$ of uniform random pixels, we have $|f(x) -  \bar{f}| \le t$ with probability at least
 $$1-\frac{Le^{-2\pi n t^2/L^2} }{\pi t \sqrt{n} }.$$
%\end{lemma}
\end{restatable}

\subsection{Capturing on-manifold behavior}
\label{sec:method}
The lemma above shows the importance of capturing the behavior of neural networks near the data manifold.  Unfortunately, the structure of image distributions is highly complex and difficult to model.  Rather than try to identify and flatten the complex structures on which images lie, we take an approach that is inspired by the recent success of the highly popular paper on the mixup regularizer \cite{zhang2017mixup},  which observed that, in addition to possessing structure near the data manifold, {\em decision boundaries are also structured in the convex hull between pairs of data points}.

We take a page from the mixup playbook and plot decision boundaries along the convex hull between data samples.  We first sample a triplet $(x_1, x_2, x_3)\sim \mathcal{D}^3$ of i.i.d.\! images from the distribution $\mathcal{D}$. Then, we construct the plane spanned by the vectors $\Vec{v_1} = x_2-x_1$, $\Vec{v_2} = x_3-x_1$ and plot the decision boundaries in this plane. To be precise, we sample inputs to the network with coordinates
\[ 
\alpha \cdot \max(\Vec{v_1} \cdot \Vec{v_1}, |\text{proj}_{\Vec{v_1}}\Vec{v_2} \cdot \Vec{v_1}|)\Vec{v_1} + \beta (\Vec{v_2} - \text{proj}_{\Vec{v_1}}\Vec{v_2})
\]
for $-0.1 \leq \alpha, \beta \leq 1.1$. 
This plotting methods using planes has several advantages. It shows the regions surrounding multiple data points at once and also the decision boundaries between their respective classes, using just one plot.  Furthermore, these classes can be chosen by the user. It also focuses on the convex hull between points rather than random directions that may point away from the manifold.

\begin{figure*}[h]
\begin{center}
\includegraphics[width=.8\linewidth]{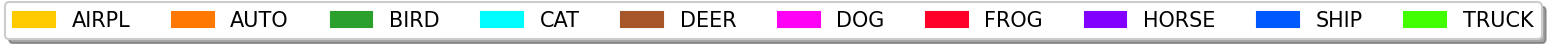}
\includegraphics[width=.9\linewidth]{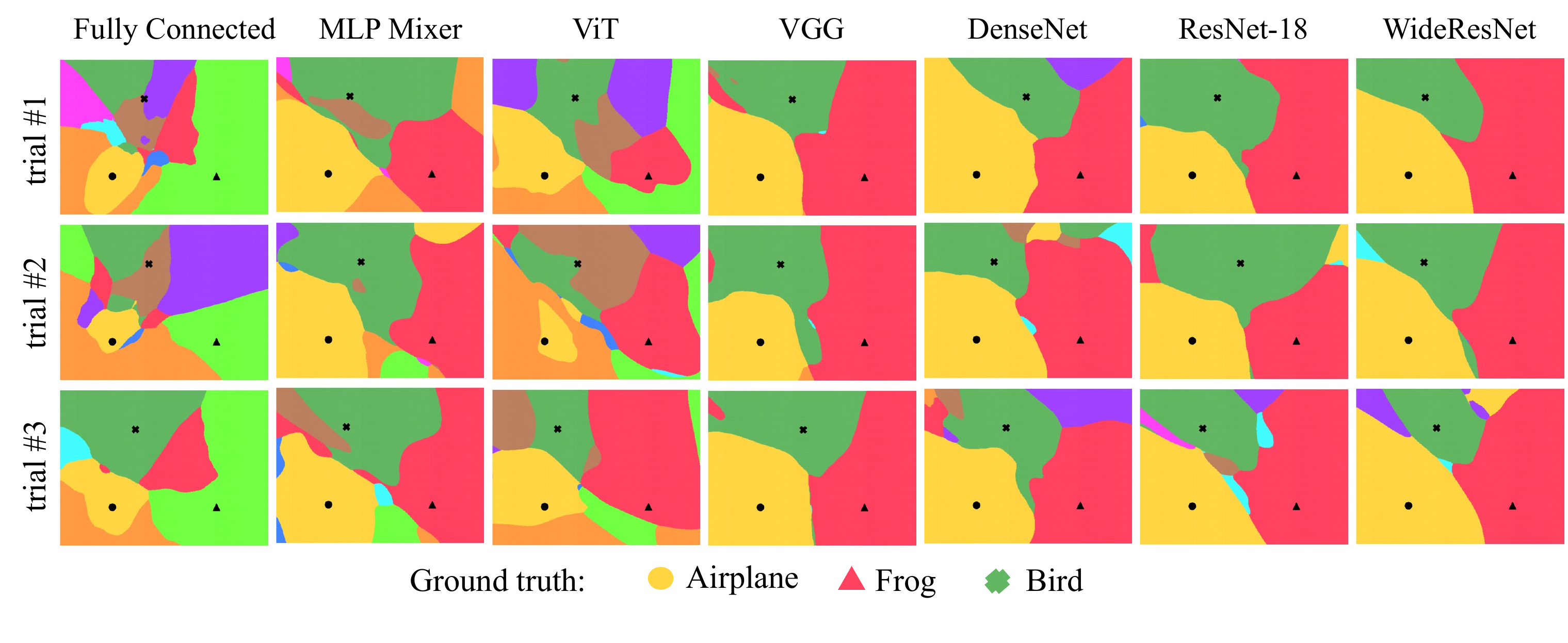}
\end{center}
\vspace{-7mm}
\caption{
Decision regions through a triplet of images, for various architectures (columns) and initialization seeds (rows).
}
\vspace{-2mm}
\label{fig:db_across_models_widths}
\end{figure*}
Figure~\ref{fig:teaser} shows decision regions plotted along the plane spanned by three data points chosen at random from the \textit{Airplane}, \textit{Frog}, and \textit{Bird} classes of CIFAR-10~\cite{krizhevsky2009learning}.
In these plots, each color represents a class label. Same color-class schema is maintained through out the paper and can be seen in legends of multiple plots.

\subsection{Experimental Setup:}
\paragraph{Architectures used:}
We select several well-known networks from diverse architecture families\footnote{Architecture implementations from \url{https://github.com/kuangliu/pytorch-cifar} and \url{https://github.com/lucidrains/vit-pytorch}}. We consider a simple Fully Connected Network with 5 hidden layers and ReLU non-linearities, DenseNet-121~\cite{huang2017densely}, ResNet-18~\cite{he2016deep}, WideResNet-28x10, WideResNet-28x20, WideResNet-28x30~\cite{zagoruyko2016wide}, ViT~\cite{dosovitskiy2020image}, MLPMixer~\cite{tolstikhin2021mlp}, and VGG-19~\cite{simonyan2014very}. For fast training, our ViT has only $6$ layers, $8$ heads, and patchsize $4$. The custom MLPMixer we use has $12$ hidden layers with hidden embedding dimension $512$ and patch size $4$. Unless otherwise stated, architectures are trained for $100$ epochs using SGD optimizer, and $3$ multi-step learning rate drops. 
Random Crop and Horizontal Flip data augmentations are used in training. 
For distillation experiments, we also use a ViT-S/16 pretrained on ImageNet~\cite{deng2009imagenet} as a teacher~\cite{rw2019timm}.
Some experiments use the Sharpness-Aware Minimization (SAM) optimizer~\cite{foret2020sharpness} adversarial radius set to $\rho=0.01$.

We select learning rates using a grid search across \{ 0.001, 0.002, 0.005, 0.01, 0.02, 0.05\} for each architecture and optimizer (Adam~\cite{kingma2014adam} and SGD) combination, and training for 200 epochs. Mean test accuracy over 3 runs per model is reported in Table \ref{tab:repro_optimizers}.

\section{Model reproducibility and inductive bias} \label{sec:repro}
It is known that neural networks can easily overfit complex datasets, and can even interpolate randomly labeled images  \cite{zhang2021understanding}.   Despite this flexibility, networks have an important {\em inductive bias} -- they have a strong tendency to converge on decision boundaries that generalize well. Our goal in this section is to display the inductive bias phenomenon using decision boundary visualizations.
We ask two questions:
\begin{itemize}[noitemsep,nolistsep]
    \item  Can a model replicate the same decision boundaries twice, given different random initializations?  
    \item Are there disparities between the inductive biases of different model families that result in different decision boundaries?
\end{itemize}
Below, we consider various sources of inductive bias, including neural architecture family, network width, and the choice of optimizer.

\subsection{Inductive bias depends on model class}
We choose three random images from the CIFAR-10 training set, construct the associated plane through input space, and plot the decision regions for 7 different architectures in 
Figure \ref{fig:db_across_models_widths}.  For each model, we run the training script three times with different random initializations.

Several interesting trends emerge in this visualization. First, we observe systematic differences between model families.  Convolutional models all share similar decision boundaries, while the boundaries of Fully Connected Nets, ViT, and MLP Mixer share noticeable differences.  For example, ViT and MLP Mixer consistently show the presence of an orange ``Automobile'' region that CNNs do not.  Fully Connected Nets show considerably more complex and fragmented decision regions than other model families.

At the same time, we observe strong reproducibility trends across runs with different random seeds. This trend is particularly high for convolutional architectures, and the effect is quite strong for WideResNet, which leads us to hypothesize that there may a link between model width and reproducibility -- an issue that we will investigate in more detail below.

\subsection{Quantitative analysis of decision regions}
\label{subsec:eval_tools}
The visualizations in Figure \ref{fig:db_across_models_widths} suggest that 
reproducibility is high within a model class, while differences in inductive bias result in low similarities across model families. %This conclusion is based only on qualitative analysis using a fairly small amount of data. 
To validate our intuitions, we use quantitative metrics derived from the decision plots averaged over many trials to provide a more sensitive and conclusive analysis.
\vspace{-3mm}\paragraph{Reproducibility Score:} We define a metric of similarity between the decision boundaries of pairs of models.  We first sample triplets $T_i = (x_0, x_1, x_2)_i$ of i.i.d. images from the training distribution. Let $S_i$ be the set of points in the plane defined by $T_i$ at which the decision regions are evaluated. We define the reproducibility score: 
\begin{equation}
\label{eq:repro}
R(\theta_1, \theta_2) = \mathbb{E}_{T_i \sim \mathcal{D}} \bigg[(|f(S_i, \theta_1) \cap f(S_i, \theta_2)|) / |S_i|\bigg]
\end{equation}
where for notation simplicity we denote the set of class predictions within each decision region as $f(S_i, \theta) = \{(x,f(x;\theta))\}_{x\in S_i} $ for a model with parameters $\theta$.
Practically, we estimate the expectation in \cref{eq:repro} by sampling $500$ triplets and $2500$ points in each truncated plane for a total of $1.25M$ forward passes. Simply put, this corresponds to the ``intersection over union" score for two decision boundary plots.

This score can quantify reproducibility of decision regions across architectures, initializations, minibatch ordering, etc. 
In earlier work~\cite{anonymous2022decision}, variability of the decision boundaries is studied by examining the similarity of predictions at test points. In contrast, our method gives a much richer picture of the variance of the classification regions not just at the input points, but also in the regions around them and can be applied to both train and test data.

\paragraph{Measuring architecture-dependent bias}

We apply the reproducibility score to measure model similiarity between different training runs with the same architecture and across different architectures.  For each model pair, we compute the reproducibility score across 5 different training runs and 500 local decision regions, each containing 2,500 sampled points (6.25M total forward passes compared). 

\begin{figure}
    \centering
    \includegraphics[width=\linewidth]{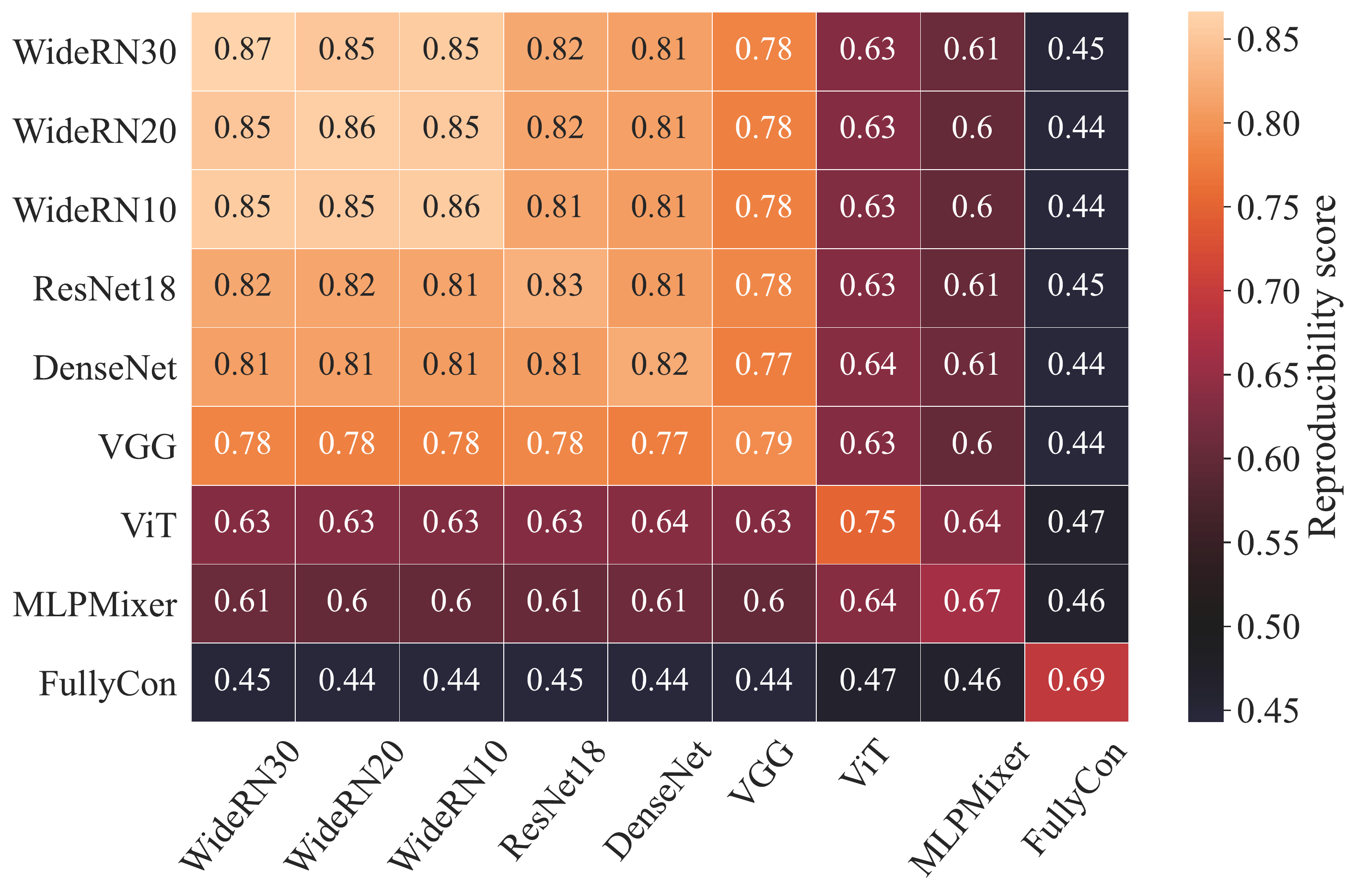}
    \caption{Reproducibility across several popular architectures.}
    \label{fig:repro_archs}
\end{figure}

Figure \ref{fig:repro_archs} shows reproducibility scores for various architectures, and we see that quantitative results strongly reflect the trends observed in the decision regions of Figure \ref{fig:db_across_models_widths}.  In particular, it becomes clear that 
\begin{itemize}[noitemsep,nolistsep]
    \item The inductive biases of all the convolutional architectures are highly similar.  Meanwhile, MLPMixer, ViT and FC models have substantially different decision regions from convolutional models and from each other.
    \item  Wider convolutional models appear to have higher reproducibility in their decision regions, with WideRN30 being both the widest and most reproducible model in this study.
    \item Skip connections have little impact on the shape of decision regions. ResNet (with residual connections across blocks), DenseNet (with many convolutional connections within blocks), and VGG (no skip connections) all share very similar decision regions.  However it is worth noting that skip connection architectures achieve slightly higher reproducibility scores than the very wide VGG network.
\end{itemize}

\subsection{Does distillation preserve decision boundaries?}
\begin{figure}
    \centering
 
    \includegraphics[width=\linewidth]{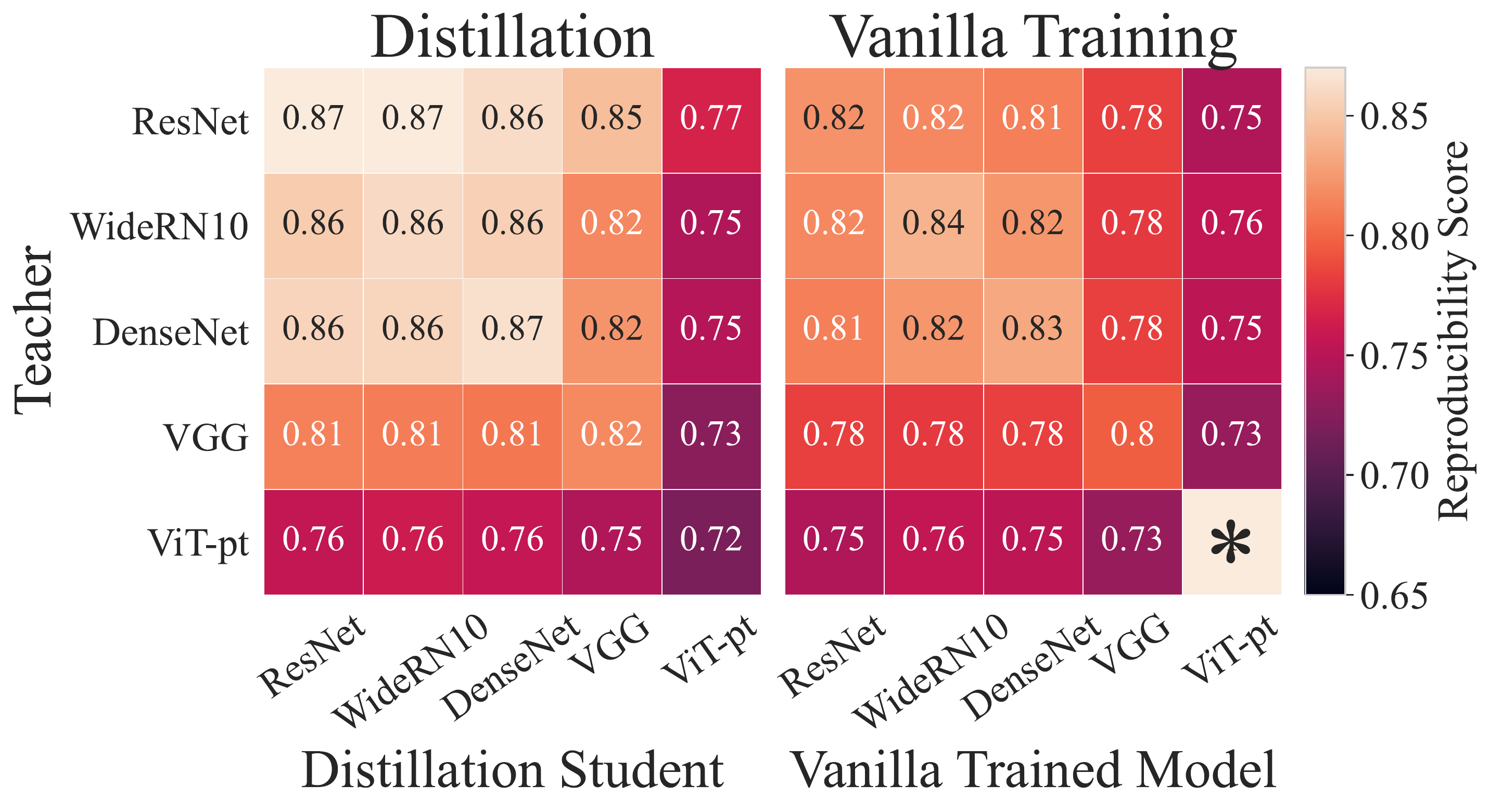}
    \caption{Differences in reproducibility comparing distilled model to vanilla trained model. *The reproducibility score is not applicable for this diagonal entry because we start from the same pre-trained model. }
    \label{fig:dist_comp}
\end{figure}

Distillation \cite{hinton2015distilling} involves training a student model on the outputs of an already trained teacher model. 
Some believe that distillation does indeed convey information about the teacher's decision boundary to the student \cite{goldblum2020adversarially}, while others argue distillation improves generalization through other mechanisms \cite{stanton2021does}. We calculate the relative similarity of the student's decision boundary to its teacher's boundary and compare this to the similarity between teacher network and a network of the student's architecture and initialization but trained in a standard fashion. Across the board, distilled students exhibit noticeably higher similarity to their teachers compared with their vanilla trained counterparts. In Figure \ref{fig:dist_comp}, we see that \emph{almost every} student-teacher combination has a higher reproducibility score than the same teacher compared to an identically initialized model trained without distillation.

\subsection{The effect of the optimizer} 
\label{subsec:optim}
In addition to the influence of initialization, data ordering, and architecture, the choice of optimizer/regularizer used during training can greatly impact the resulting model \cite{geiping2021stochastic}. Thus, we study the effect of optimizer choice on the reproducibility of a network's decision boundary. In Table \ref{tab:repro_optimizers}, we can see that SAM~\cite{foret2020sharpness} induces more reproducible decision boundaries than standard optimizers such as SGD and Adam. This observation suggests that SAM has a stronger regularization effect. However, more regularization doesn't always mean better test accuracy. For example, for MLPMixer and ViT, using SAM does not always achieve the highest test accuracy but does achieve the highest reproducibility. 

\begin{table}[]
    \centering
    \begin{tabular}{cccc}
    \multicolumn{4}{c}{Reproducibility} \\
        \toprule
						& Adam &	SGD	 & SGD + SAM \\
                        \midrule
    ResNet-18	&	79.81\%	&	83.74\% &	\textbf{87.22\% }\\
    VGG	& 81.19\%	& 80.92\% &	\textbf{84.21\% }\\
    MLPMixer	& 67.80\% &	66.51\%	& \textbf{68.06\%} \\
    VIT	& 69.55\% &	75.13\%	 & \textbf{75.19\%} \\
    \bottomrule   \vspace{-2mm}\\
    \multicolumn{4}{c}{Test Accuracy} \\
        \toprule
		& Adam &	SGD	 & SGD + SAM \\
		\midrule
	    ResNet-18 & 93.04 & 95.30 & \textbf{95.68} \\
        VGG	& 92.87 &	93.13 &	\textbf{93.90} \\
        MLPMixer & \textbf{82.22} & 	82.04 & 82.18\\
        VIT	& 70.89 &	\textbf{75.49} &	74.72 \\
    \bottomrule
    \end{tabular}
    \caption{Reproducibility of different models when using different optimizers. SGD produces more reproducible decision boundaries relative to Adam, and SGD+SAM almost always consistently increase reproducibility of the model relative to SGD.}
    \vspace{-2mm}
    \label{tab:repro_optimizers}
\end{table}

\section{Double descent}
\label{sec:double_descent}
\begin{figure}[t]
\begin{center}
\includegraphics[width=\linewidth]{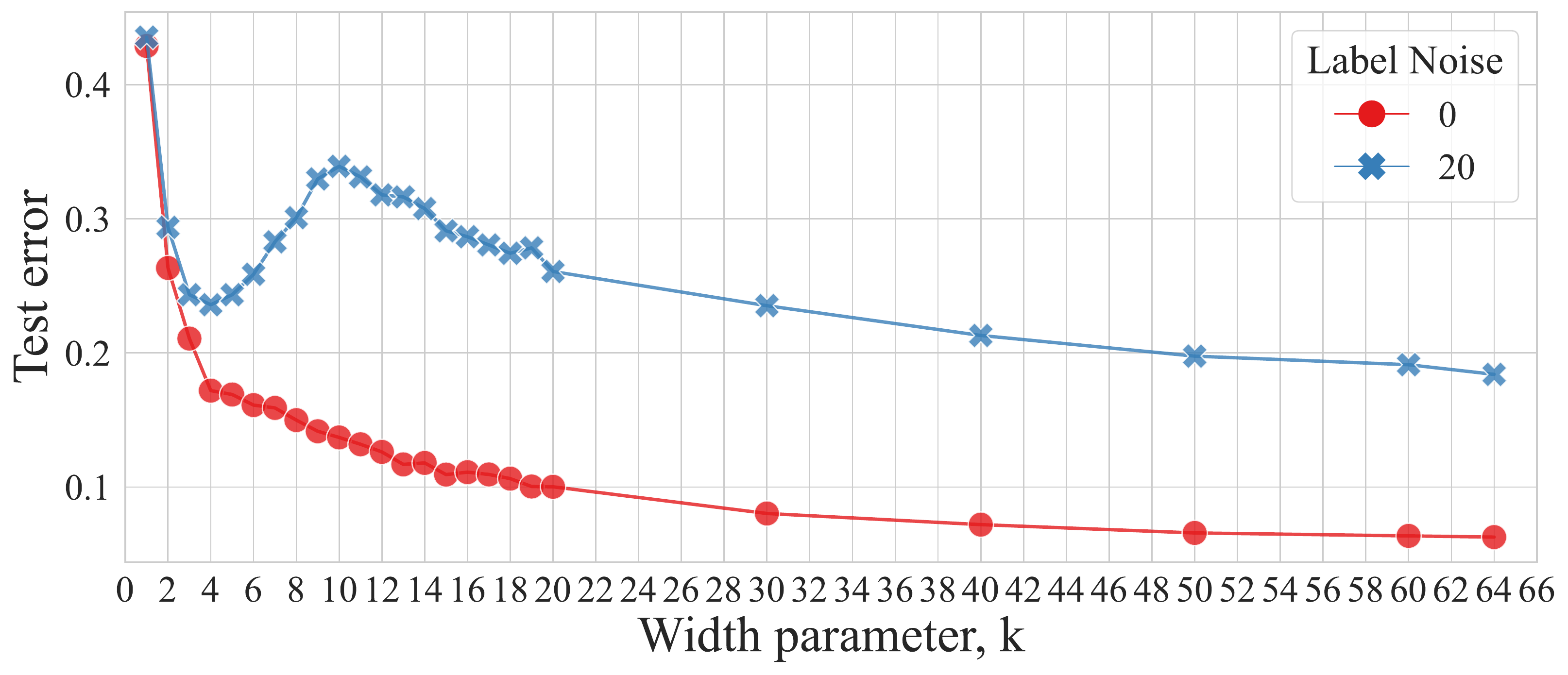}
\end{center}
\vspace{-6mm}
\caption{
Test error curves with 0 and 20\% label noise in training.
}
\vspace{-2mm}
\label{fig:dd_test_error}
\end{figure}

In classical learning theory,  it is thought that models with too few parameters (e.g., low width) generalize poorly because they are not expressive enough to fit the data, while models with too many parameters generalize poorly because of over-fitting.  This is known as the Bias-Variance trade-off~\cite{geman1992neural}.  In contrast, the strong inductive bias of neural networks enables them to achieve good performance even with extremely large numbers of parameters.
Belkin et al.~\cite{belkin2019reconciling} and Nakkiran et al.~\cite{nakkiran2019deep} have shown that under the right training conditions, we can see neural models operating in both the classical and over-parameterized regimes.  This is depicted in Fig. \ref{fig:dd_test_error}, which plots test error as a function of model width on CIFAR-10.  We observe a classic U-shaped curve for widths less than 10~(the underparametrized regime). For models of width greater than 10, the test error fall asymptotically (overparametrized regime). This behaviour is referred to as ``double descent'' and discussed in generality in Belkin et al.~\cite{belkin2019reconciling}. 
Between the two regimes is a model that lives at the ``interpolation threshold''; here, the model has too many parameters to benefit from classical simplicity bias, but too few parameters to be regularized by the inductive bias of the over-parameterized regime. 
Double descent has been studied rigorously for several simple and classical model families, including kernel methods, linear models, and simple MLPs~\cite{advani2020high, opper1995statistical, opper2001learning, spigler2018jamming, hastie2019surprises,muthukumar2020harmless, dar2021farewell}. 
Double descent is now well described for linear models and random feature networks in~\cite{hastie2019surprises,d2020double,dar2021farewell,adlam2020understanding}. In the classical regime, bias decreases with increased model complexity, while the variance increases at the same time, resulting in a U-shaped curve. Then, in the overparameterized regime, the variance decreases rapidly while bias remains low \cite{neal2018modern,yang2020rethinking}.  In our studies above, we visualized the over-parameterized regime and saw that models become highly reproducible, with wide architectures producing nearly identical models across training runs.  These visualizations captured the low-variance of the over-parameterized regime.

In this section, our goal is to gain insight into the model behaviors that emerge at the interpolation threshold, causing double descent. We observe closely what is happening at critical points~(i.e., the transition between the under and overparameterized regimes), and how the class boundaries transition as we increase the capacity of the model class. We find that the behaviour of class boundaries aligns with the bias-variance decomposition findings of~\cite{neal2018modern,yang2020rethinking}, however the model instabilities that cause variance to spike in neural networks is manifested as a complex fragmentation of decision space that is not, to the best of our knowledge, described in the literature on classical models.

\paragraph{Experimental setup:} We follow the experimental setting from Nakkiran et al.~\cite{nakkiran2019deep} to replicate the double descent phenomenon for ResNet-18~\cite{he2016deep}. We increase model capacity by varying the number of filters in the convolutional layers by a ``width'' parameter, $k$. Note that a standard ResNet-18 model has $k=64$ and lives in the over-parameterized regime on CIFAR-10. We train models with cross-entropy loss and the Adam optimizer with learning-rate 0.0001 for 4000 epochs. This gentle but long training regiment ensures stability and convergence for the wide range of models needed for this study. 

It was observed in~\cite{nakkiran2019deep} that label noise is important for creating easily observable double descent in realistic models. We train two sets of models, one with a clean training set and another with $20\%$ label noise (uniform random incorrect class labels).  In both cases, we use the standard (clean) test set. %To add label noise, we randomly sample 20\% of the training data and uniformly random incorrect labels are assigned to them 
For noisy experiments, the same label errors are used across epochs and experiments. RandomCrop and RandomHorizontalFlip augmentations are used while training. 
We observe a pronounced double-descent when label noise is present. See Figure~\ref{fig:dd_test_error}, which replicates the double-descent curve of Nakkiran et al.~\cite{nakkiran2019deep}. 

We focus on several important model widths: $k=4$ is the local minimum of test error in the underparametrized regime, and $k=10$ achieves peak error ($\approx$ interpolation threshold) beyond which the test error will continually fall. 
We refer the reader to Appendix~\ref{appendix_sec:double_descent_results} for training error plots showing the onset of interpolation near $k=10.$

\begin{figure*}
     \centering
     \begin{subfigure}[b]{.99\linewidth}
         \centering
         \includegraphics[width=.9\linewidth]{fig/dd/db_plots/legend_class.png}
\includegraphics[width=0.9\linewidth]{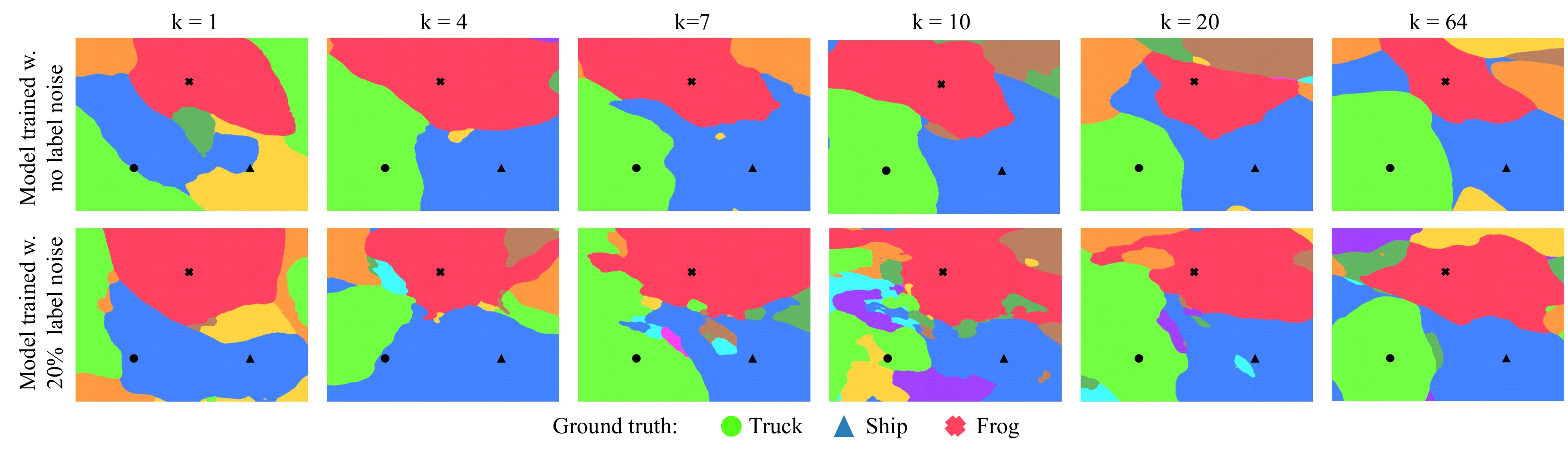}

         \caption{All the points in the triple are from different classes, and are correctly labeled in the train set (even in the label noise case).}
         \label{fig:dd_3random}
     \end{subfigure}
     \vfill
     \begin{subfigure}[b]{.99\linewidth}
         \centering
\includegraphics[width=.9\linewidth]{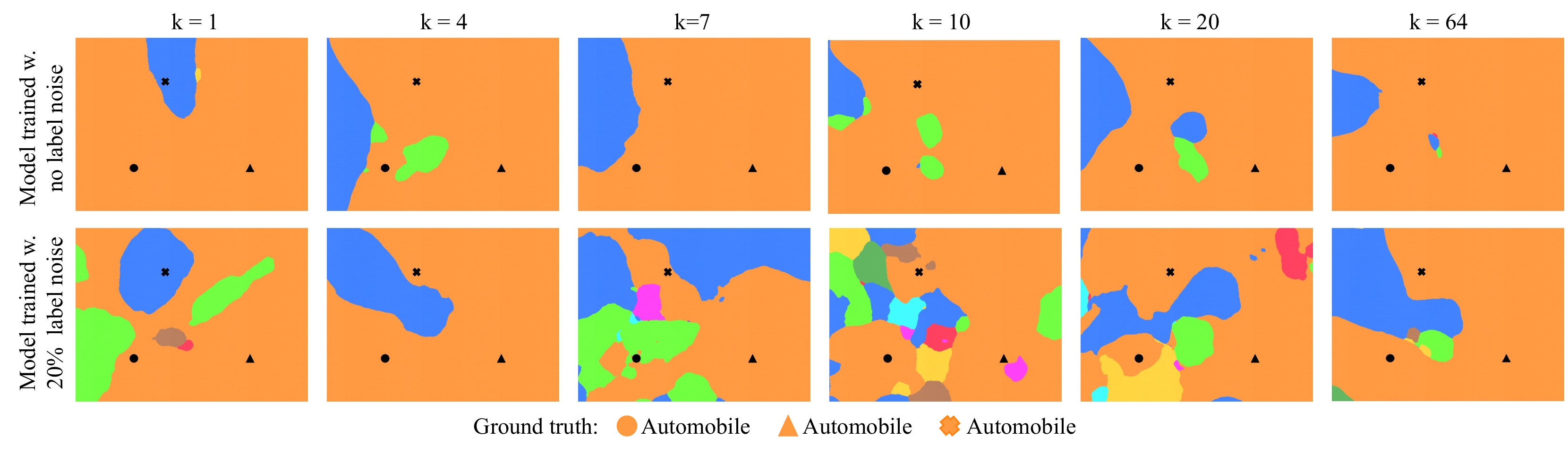}

         \caption{All points in the triple are from the same class, Automobile, and are correctly labeled in the train set (even in the label noise case).}
         \label{fig:dd_3corr}
     \end{subfigure}
     
\caption{\textbf{Decision boundaries for models of varying width.} Label noise induces chaotic fragmentation of decision regions as we cross the threshold of interpolation (k=10), while very narrow and wide models remain smooth. 
\vspace{-4mm}% and smooth/stable regions in the under- and over-parameterized regimes.
}
\label{fig:dd_dbs_all}

\end{figure*}

\subsection{How do decision boundaries change as we cross the interpolation threshold?}

In Figure \ref{fig:dd_dbs_all}, we plot decision boundaries for models trained with and without label noise and with varying capacities. As above, visualizations take place in the plane spanned by three data points.  We present examples using two different methods for sampling -- one with all three images from the same class and one with three different classes. The three images are drawn from the training set and are correctly labeled (even for the experiments involving label noise). Similar behaviours are observed for other randomly sampled images and with other combinations of classes.  See Appendix~\ref{appendix_sec:double_descent_results} for additional examples.

As we move from left to right in the figure, model capacity sweeps from $k=1$ (under-parameterized) to $k=64$ (standard ResNet-18, which is over-parameterized). As the models become increasingly over-parameterized, the models are getting confident about their predictions, as seen by the intensity of the color. When models are trained with clean labels, the model fits all three points with high confidence by the time $k=4,$ and the decision boundaries change little beyond this point. 

The mechanism behind the error spike in the double descent curve is captured by the visualizations using label noise. In this case, the under-fitting behavior of the classical regime is apparent at $k=4$, as the model fits only 1 out of 3 points correctly, and confidence in predictions is low. When we reached $k=10$ (the interpolation threshold), the model fits most of the training data, including the three points in the visualization plane. As we cross this threshold, the decision regions become chaotic and fragmented. By the time we reach $k=20$, the fragmentation is reduced and class boundaries become smooth as we enter the over-parameterized regime.

\begin{figure}[h]
\begin{center}
\includegraphics[width=0.95\linewidth]{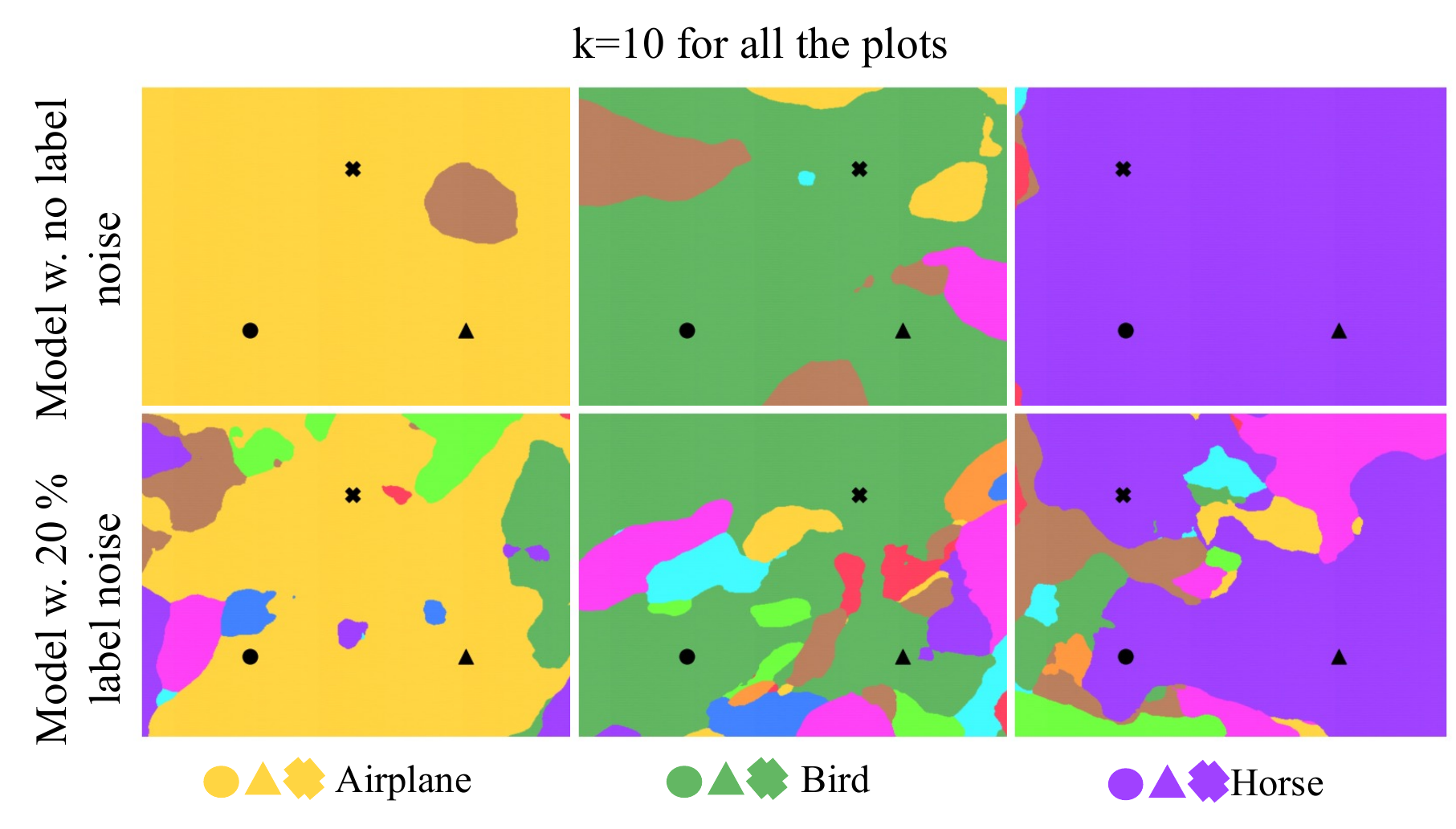}
\end{center}
\vspace{-6mm}
\caption{
Decision boundaries of 3 correctly labeled points at $k=10$ on models with and without label noise.
}
\vspace{-3mm}
\label{fig:dd_3corr_10k}
\end{figure}

To refine our picture of double descent, we visualize the class boundaries at $k=10$ for a range of different image triplets in Figure~\ref{fig:dd_3corr_10k}, both with and without label noise.  We see that in the label noise case, where double descent is observed, there is a clear instability in the classification behavior at the interpolation threshold.

\begin{figure}
\begin{center}
% \begin{overpic} 
% [width=\linewidth]
% {example-image-a}
% \end{overpic}
\includegraphics[width=0.95\linewidth]{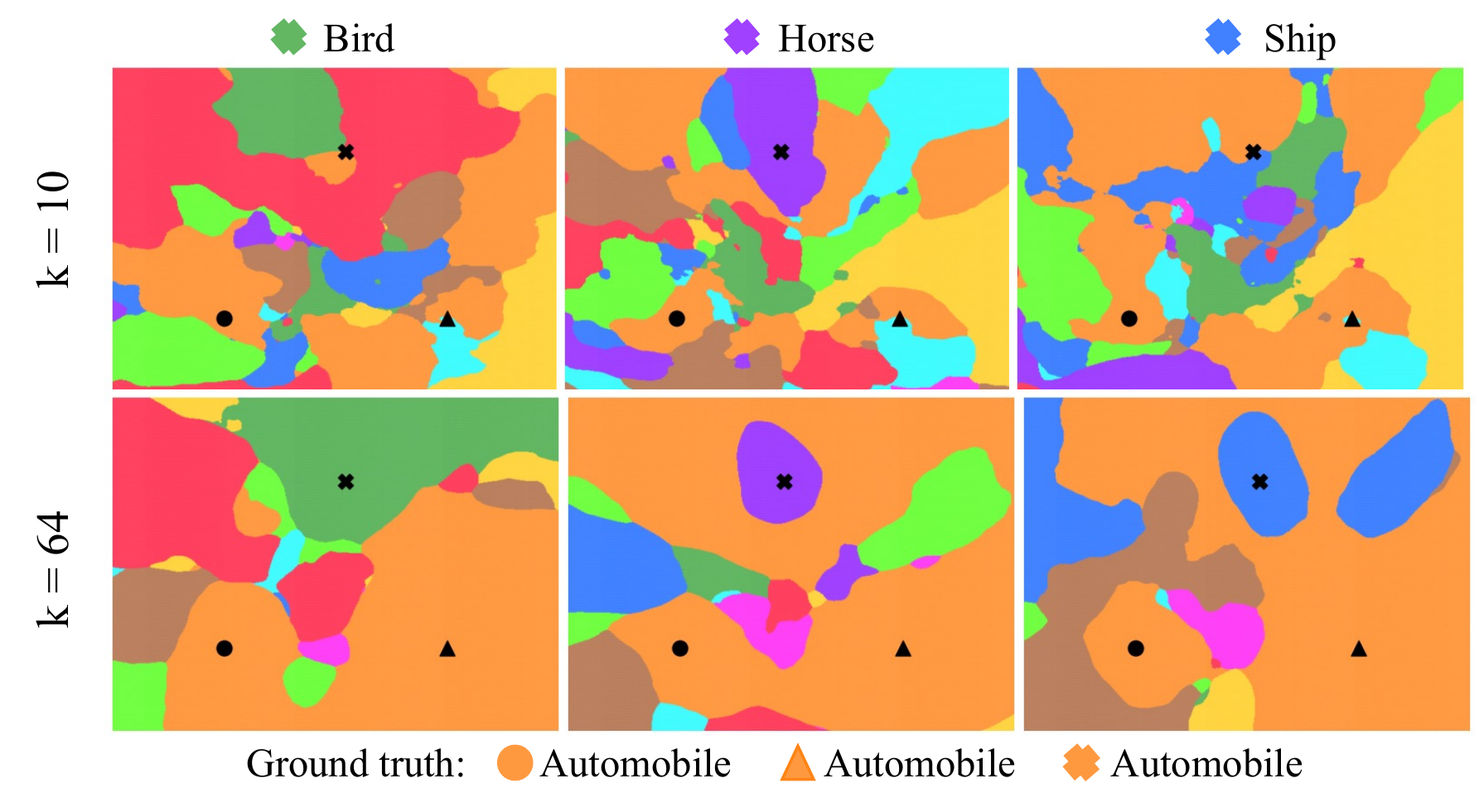}
\end{center}
\vspace{-6mm}
\caption{
Decision boundaries with 1 mislabeled automobile and 2 correctly labeled automobiles. Each column represents a different image triplet. The mislabeled point is marked by $\mathbf{x}$. 
}
\vspace{-3mm}
\label{fig:dd_mislabs_10v64}
\end{figure}

Let's now see what happens to the decision boundaries around mislabeled images.  Figure~\ref{fig:dd_mislabs_10v64} shows decision boundaries around three points from the \textit{Automobile} class, where one of the points is mislabeled in the training set. When $k=10$, we see chaotic boundaries.  The mislabeled points are assigned their (incorrect) dataset label, but they are just barely interpolated in the sense that they lie very near the decision boundary.  For $k=64$, the boundaries are seemingly regularized by inductive bias; the mislabeled points lie in the center of their respective regions, and boundaries are much more smooth.

Having observed the qualitative behaviour of correctly labeled and mislabeled points in models with and without label noise at various capacities, we ask the following questions:
\begin{itemize}[noitemsep,nolistsep]
    \item Can quantitative methods validate that fragmentation behavior persists across multiple decision regions at the interpolation threshold and vanishes elsewhere? 
    \item Is the fragmentation at the interpolation threshold indeed caused by model variance?  In other words, do we observe different decision boundaries across training runs, or are the chaotic regions reproducible like the regions we observed in the over-parameterized regime?
    \item What is the mechanism for the decrease in test error in the wide model regime? Is it caused by shrinkage of the misclassified regions around mislabeled points, causing them to stop contaminating the test accuracy? Or is it merely caused by the vanishing of unnecessary fragmentation behavior for large $k$?
\end{itemize}
In the subsequent sub-sections, we investigate these issues using quantitative measurements of decision regions.

\subsection{Quantifying fragmentation}
We have observed that decision regions appear to become highly fragmented as we cross the interpolation threshold.
To verify that our results are repeatable across many experiments and triples, we introduce the \emph{fragmentation score}, which counts the number of connected class regions in the plane spanned by a triplet of images.

Let $S_i$ be a local classification region spanned by a triplet $T_i$. We create a decomposition $S_i(\theta) = {\cup}_{j=1}^{n_i} P_j(\theta)$ where each $P_j(\theta)$ is a disjoint, maximal, path-connected component corresponding to a \emph{single} predicted class label for the model with parameters $\theta$. %So, for any $x_1, x_2 \in P_j$ there exists a continuous path $g:[0,1] \rightarrow S_i$ with $g(0) = x_1$, $g(1) = x_2$, and $f(g(x), \theta) = y_j$ $\forall x \in [0,1]$ 
The fragmentation score $F(\theta, T_i)$ of model $\theta$ within the decision region defined by $T_i$ is then the number of path-connected regions.  The overall fragmentation score for a model is  
\begin{align}
\label{eq:fragmentation}
%F(\theta, T_i) = |{\{P_j\}}_{j=1}^{n_i}| = {n_i} \\
F(\theta) = \mathbb{E}_{T_i \sim \mathcal{D}}\, F(\theta, T_i).
\end{align}
In practice,
we compute the fragmentation score of a model using a watershed method to find connected regions in decision region spanned by the triplet and then by averaging such fragmentation counts over 1000 triplets.

Note that prior work~\cite{fawzi2018empirical} proposes a metric to understand class connectivity that requires solving a non-convex optimization problem to find an explicit path between any two given points. In contrast, our fragmentation score is scalable, does not require any backward passes to approximate the complexity of decision boundaries, and can be averaged over a large number of input triples.

\begin{figure}[t]
\begin{center}
% \begin{overpic} 
% [width=\linewidth]
% {example-image-a}
% \end{overpic}
\includegraphics[width=\linewidth]{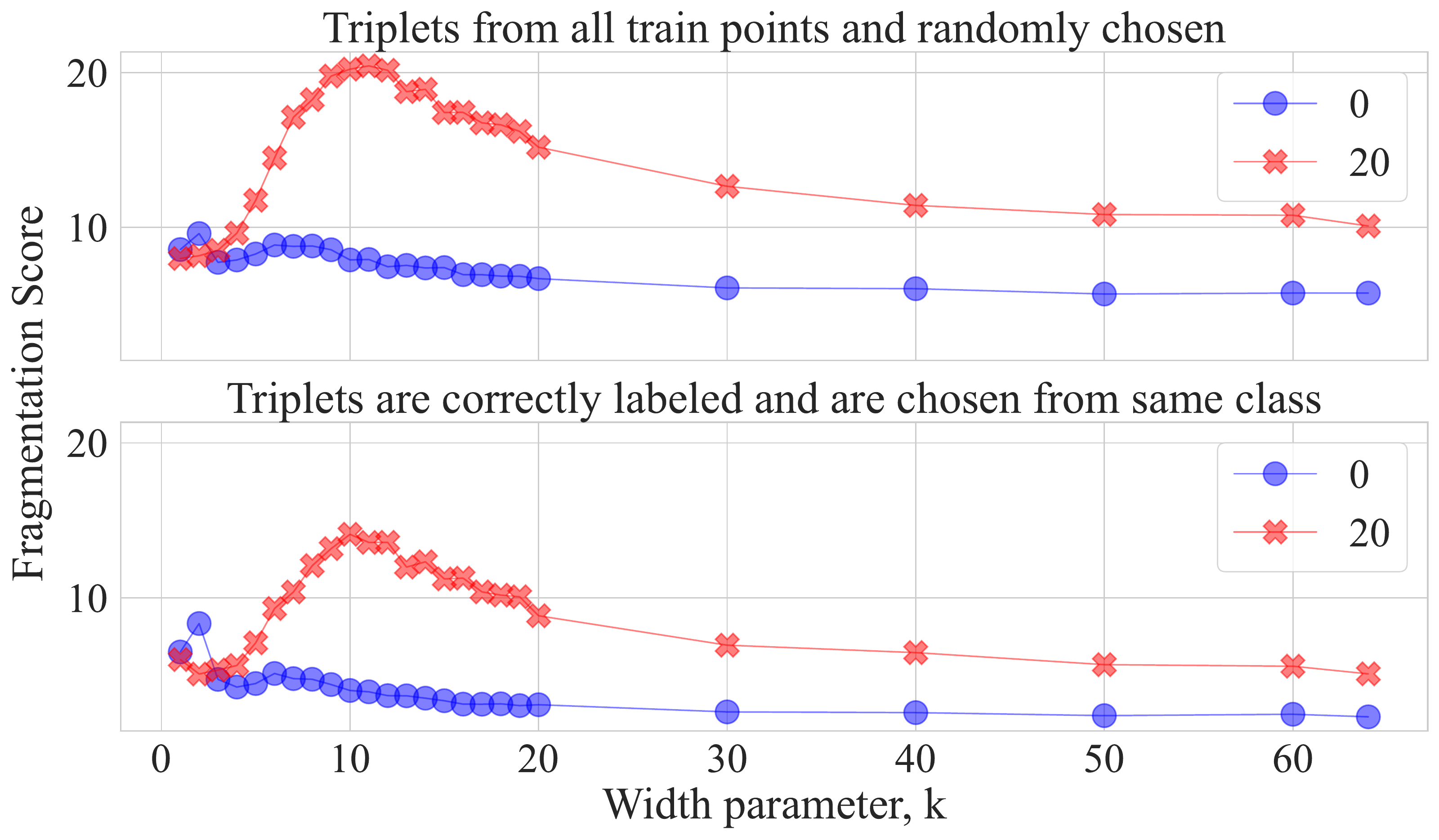}
\end{center}
\vspace{-6mm}
\caption{
Fragmentation scores as a function of model width for models trained with and without label noise.}
\label{fig:dd_fragmentation}
\end{figure}

Fragmentation scores as a function of model width are depicted in Figure \ref{fig:dd_fragmentation}. With label noise, we see a sharp peak in fragmentation score as the model capacity crosses the interpolation threshold, confirming our observations from the visualization in the figures above.  Interestingly, this highly sensitive analysis is also able to detect a peak (around $k=7$) in the fragmentation score for models trained without label noise.  The bottom part of Figure~\ref{fig:dd_fragmentation} quantifies the fragmentation trend for the decision regions spanned by triplets of the same class (like in Figure~\ref{fig:dd_3corr_10k}).

\iffalse
When we calculate the fragmentation score on triplets from the same class, it can help us understand how connected the points are. Prior work Fawzi et al~\cite{fawzi2018empirical} proposed a metric to understand the class connectivity, but to compute that metric, one has to solve optimization problem to find path between any two given points. Our fragmentation score is scalable quite easily and do not need any backward passes in computation and provides good approximation to understand the complexity of decision boundaries.
\fi

\subsection{Quantifying class region stability} \label{sec:fragstable}

\begin{figure}[h]
     \centering
     \begin{subfigure}[b]{\linewidth}
         \centering
         \includegraphics[width=\linewidth]{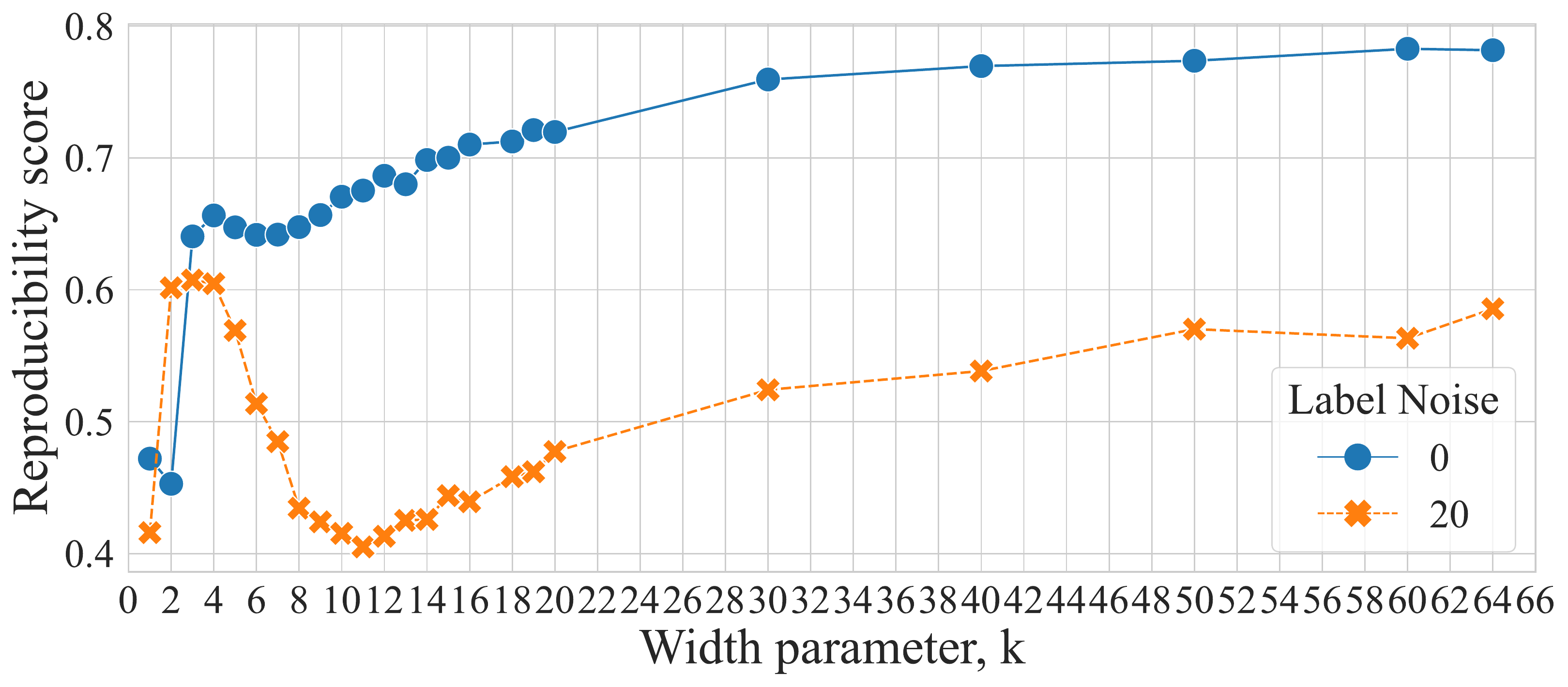}
        %  \caption{3 random images}
        %  \label{fig:dd_3random}
     \end{subfigure}
    \vspace{-6mm}
\caption{Reproducibility with respect to random initialization for models of different widths.}
\label{fig:dd_repro_all}
\end{figure}
 \begin{figure}[t]
\begin{center}
\includegraphics[width=\linewidth]{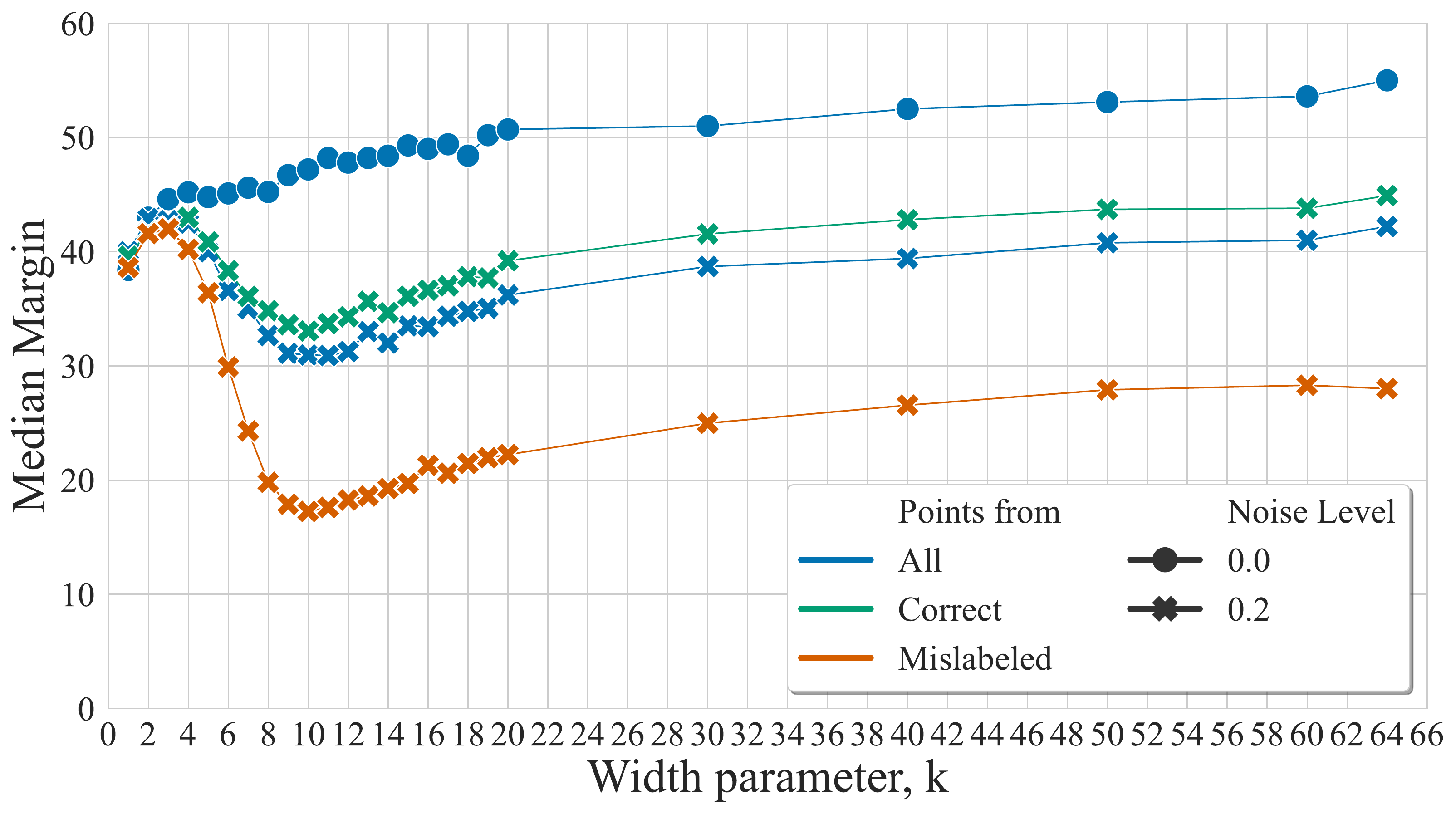}
\end{center}
\vspace{-6mm}
\caption{
Median Margins - models with and without label noise. Y-axis reflects the average perturbation size needed to reach decision boundary in a random direction.}
\label{fig:dd_margins}
\end{figure}

Theoretical studies of double descent predict that, for simple linear model classes, model variance spikes near the interpolation threshold as decision regions become highly unstable with respect to noise in the data sampling process. Using reproducibility scores, we observe that the fragmentation of neural decision boundaries at the interpolation threshold is associated with high variance and model instability. Figure~\ref{fig:dd_repro_all} shows reproducibility scores across model capacities with and without label noise.  We see that reproducibility across training runs is high in the under- and over-parameterized regimes, but breaks down at the interpolation threshold.  Interestingly, our quantifications are sensitive enough to detect a dip in reproducibility even without label noise, although the variance introduced by this effect is not strong enough to cause double descent.  Note that the model variance in Figure~\ref{fig:dd_repro_all}  is caused by differences in random initialization.  Classical convex learning theory studies variance with respect to random data sampling.  We find that a similar curve is produced by freezing initialization and randomizing the sampling process (see Appendix~\ref{appendix_sec:double_descent_results}).

\subsection{Why does label noise amplify double descent?}

The dramatic effect of label noise near the interpolation threshold could be caused by two factors:  (i) the necessary regions of incorrect class labels that must emerge around mislabeled points for the model to interpolate them, or (ii) instability in the class boundaries, resulting in oscillations that are not needed to interpolate the data.  Quantitative evidence presented above suggests that (ii) is the predominant mechanism of double descent. The lower fragmentation scores in over-parameterized regime (where almost all mislabeled points are interpolated) compared to critical regime as seen in Figure~\ref{fig:dd_fragmentation} shows that the extra regions are not needed for interpolation.

To lend more strength to this conclusion, we investigate hypothesis (i) by measuring the ``mean margin,'' which we define to be the average distance between an image and the edge of its class region in a random direction. For each image, we approximate this value using a bisection search in 10 random directions. We compute the mean margin for 5000 data points and report the median for models with and without label noise in Figure~\ref{fig:dd_margins}. 

Both with and without label noise, the margins are increasing for $k \ge 10$ (the over-parameterized regime). The interesting observation is, when we computed margins of only the mislabeled points, they go up too! The fact that test error descends, even as the regions around mislabeled points grow, lends further strength to the notion that double descent is predominantly driven by the ``unnecessary'' oscillations resulting from model instability, and not by the error bubbles around mislabeled points.

\section{Conclusion}
In this article, we use visualizations and quantitative methods to investigate model reproducibility, inductive bias, and double descent from an empirical/scientific perspective. These explorations reveal several interesting behaviors of neural models that we do not think have been previously observed. Curiously, the results of Section \ref{sec:repro} indicate that different model families achieve low test error by different inductive strategies; While ResNet-18 and ViT make similar predictions on test data, there are dramatic differences in the decision boundaries they draw.  Also, while our studies of double descent found that the model instability predicted for linear models is also observed for neural networks, we saw that this instability is manifested as the dramatic fragmentation of class regions.  These oscillations in the model output are reminiscent of ``Gibbs phenomenon,'' and do not appear to be described in the theoretical literature.

\section{Acknowledgements}
This work was supported by the ONR MURI program, the Office of Naval Research, the National Science Foundation (DMS-1912866), and DARPA GARD (HR00112020007).  Additional funding was provided by Capital One Bank and Kulkarni Summer Research Fellowship.
%%%%%%%%% REFERENCES
\bibliographystyle{ieee_fullname}
\bibliography{main}
\newpage
\appendix

% --- PDF will be split by an editor (e.g. macOS preview), so need to restart from page 1
\setcounter{page}{1}

% --- repeat the title (AT: haven't found a more elegant way to do this...)
\twocolumn[
\centering
\Large
\textbf{Can You Learn the Same Model Twice? Investigating Reproducibility and Double Descent from the Decision Boundary Perspective} \\
\vspace{0.5em}Supplementary Material \\
\vspace{1.0em}
] %< twocolumn
\appendix
\section{Proof of Lemma 2.1}
\label{appendix_sec:lemma_proof}

For clarity, we restate the lemma here.
\mainlemma*

Consider a set $\mathcal{A}\subset [0,1]^n,$ and let $d$ denote the $\ell_2$ distance metric. We define the $\epsilon$-expansion of the set $\mathcal{A}$ as $\mathcal{A}(\epsilon) = \{x\in [0,1]^n\, |\,\, d(x,\mathcal{A}) \le \epsilon\}$. In plain words, $\mathcal{A}(\epsilon)$ is the set of all points lying within $\epsilon$ units of the set $\mathcal{A}$. 

Our proof will make use of the isoperimetric inequality first presented by Ledoux \cite{ledoux2001concentration}. We use the following variant with tighter constants proved by Shafahi et al.\! in \cite{shafahi2018adversarial}.  
\begin{lemma}[Isoperimetric inequality on the unit cube]
 \label{cubeiso}
Consider a measurable subset of the cube $\mathcal{A} \subset [0,1]^n,$ and a 2-norm distance metric $d(x,y)=\|x-y\|_2$. Let $ \Phi(z)=(2\pi)^{-\frac{1}{2}}\int_{-\infty}^z e^{-t^2/2} dt,$ and let $\alpha$ be the scalar that satisfies $\Phi(\alpha) =\text{vol}[\mathcal{A} ].$ Then
\begin{align} \label{tightPhiBound}
\text{vol}[\mathcal{A}(\epsilon)]  \ge \Phi\left( \alpha+ \epsilon\sqrt{2\pi }\right).
\end{align}
In particular, if $\text{vol}(\mathcal{A})\ge 1/2,$ then we simply have 
\begin{align}\label{cleanCubeBound}
\text{vol}[\mathcal{A} (\epsilon)]  \ge 1- \frac{e^{-2\pi  \epsilon^2  }}{2\pi \epsilon }.
\end{align}
\end{lemma}

To prove Lemma 2.1, we start by choosing $\mathcal{A} = \{x| f(x) \le \bar{f}\}$. 
Now, consider any $x\in \mathcal{A} \left(t\frac{\sqrt{n}}{L}\right).$ From the Lipschitz bound on $f$ we have 
 $$|f(x) - f(y)| \le \frac{L}{\sqrt{n}}\|x-y\|,$$
 for any $y$.  If we choose $y=\arg\min_{z\in \mathcal{A}} \|x-z\|$ to be the closest point to $x$ in the set $\mathcal{A}, $  we have that $\|z-y\|\le t\frac{\sqrt{n}}{L},$ and so
  $$|f(x) - f(y)| \le t.$$
 But $f(y)\le \bar{f}$ because  $y\in \mathcal{A}.$ From this, we see that for any choice of $x\in \mathcal{A} \left(t\frac{\sqrt{n}}{L}\right)$ we have
 \begin{align} \label{eq:le}
     f(x) - \bar{f} \le  t.
      \end{align}
Recall that $\bar{f}$ is the median value of $f$ on the unit cube, and so we have that 
 $$\text{vol}[\mathcal{A}] \ge \frac{1}{2}.$$ 
We can then apply Lemma \ref{cubeiso} with $\epsilon=t\frac{\sqrt{n}}{L}$, and we see that 
   $$\text{vol}\left[\mathcal{A} \left(t\frac{\sqrt{n}}{L}\right)\right]  \ge 1- \frac{Le^{-2\pi t^2 n/L^2}}{2\pi t \sqrt{n}}.$$
 We conclude that a {\em randomly} chosen $x\in [0,1]^n$ will lie in $\mathcal{A} \left(\frac{t\sqrt{n}}{L}\right),$ and therefore satisfy \eqref{eq:le} with probability at least $1- \frac{Le^{-2\pi t^2 n/L^2  }}{2\pi t \sqrt{n}}.$
 
 An analogous argument with $\mathcal{A} = \{x| f(x) \ge \bar{f}\}$ shows that a randomly chosen $x\in [0,1]^n$ will satisfy 
 \begin{align} \label{eq:ge}
     \bar{f}-f(x) \le  t.
      \end{align}
      with the same probability.
     Applying a union bound, we see that a randomly chosen $x$ will satisfy \eqref{eq:le} and \eqref{eq:ge} simultaneously with probability at least
      $1- \frac{Le^{-\pi t^2 n/L^2 }}{\pi t \sqrt{n}}.$

\section{Decision regions}
\label{appendix_sec:offmanifold_dbs}
\begin{figure*}[t]
\begin{center}
\includegraphics[width=.8\linewidth]{fig/dd/db_plots/legend_class.png}
\includegraphics[width=.8\linewidth]{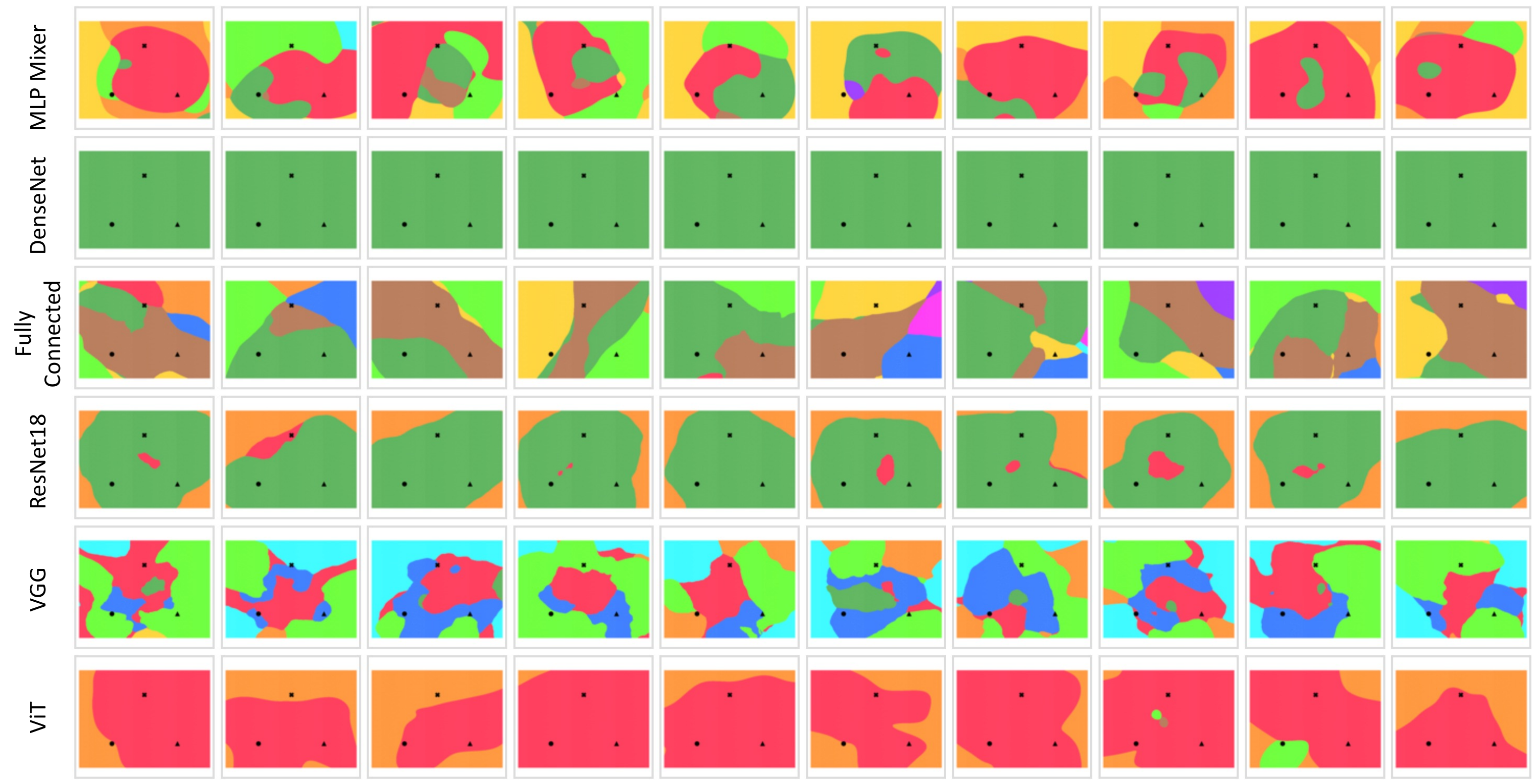}
\end{center}
\caption{Decision regions when all the images are uniformly sampled. Each row corresponds to a model,while each column is a new sampling of the triplet}
\label{appendix_fig:db_uniform}
\end{figure*}

\begin{figure*}[t]
\begin{center}
\includegraphics[width=.8\linewidth]{fig/dd/db_plots/legend_class.png}
\includegraphics[width=.8\linewidth]{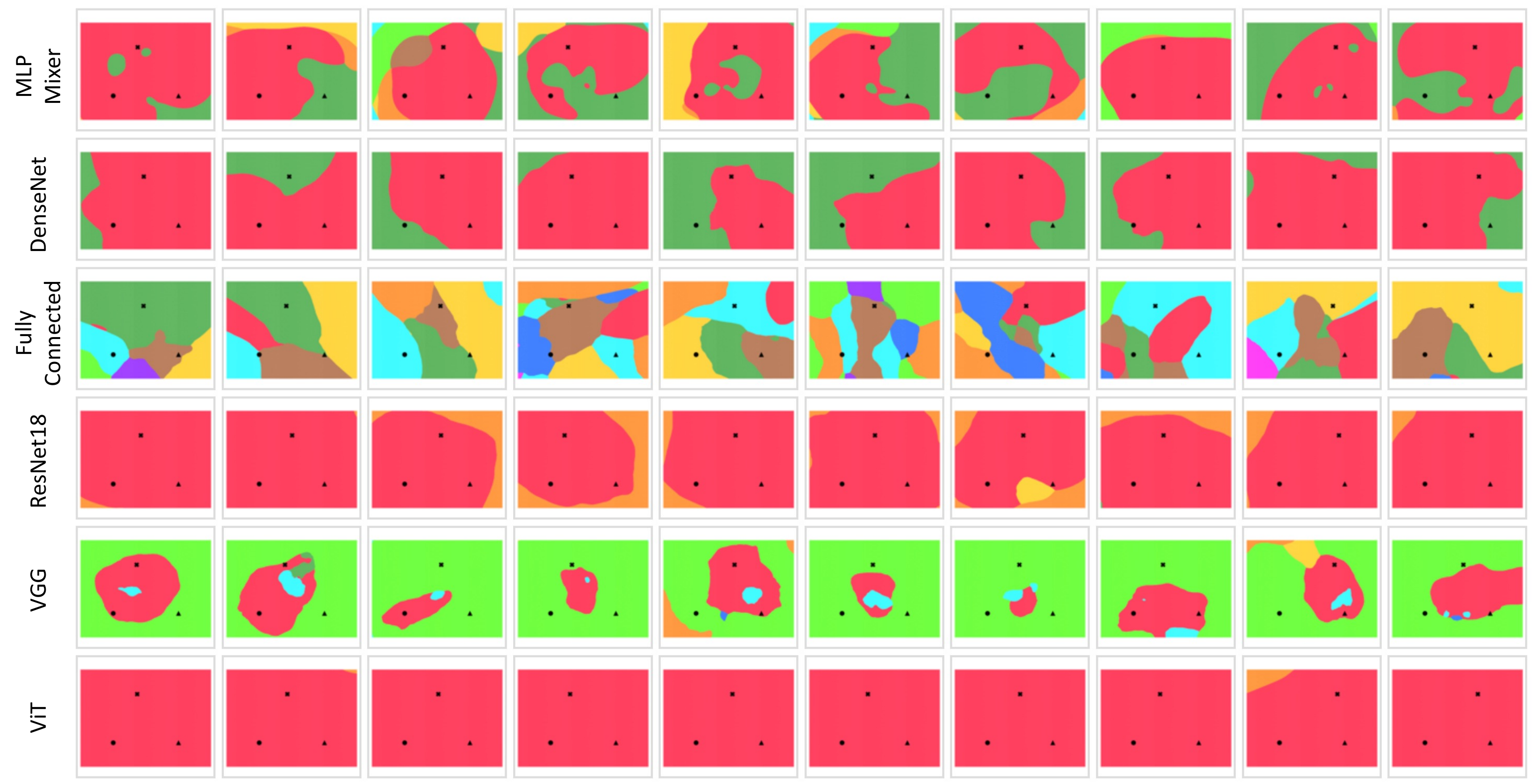}
\end{center}
\caption{Decision regions when the pixels are randomly shuffled. Each row corresponds to a model,while each column is a new sampling of the triplet. Extended version of Fig 2.}
\label{appendix_fig:db_shuffled}
\end{figure*}

\paragraph{Off-manifold decision regions}
We present a few off-manifold decision boundaries in this section. In Fig.~\ref{appendix_fig:db_uniform}, we show decision regions of multiple off manifold images where all the pixels are uniformly sampled in the image space. Each row is a model, and each column is a randomly sampled triplet. We observe that the decision regions assigned to such off-manifold images are quite uniform for a given model. For example, in DenseNet, all such images are assigned to \textit{Bird} class, while in ViT, they are assigned to \textit{Frog} or \textit{Automobile}. In Fig.~\ref{appendix_fig:db_shuffled}, we show decision regions for a multiple triplets of shuffled images. (Expanded version of Fig.2). Even in this type of off-manifold images, we see a similar pattern that the models are assigning the samplings to a certain set of classes. This emphasises that the decision regions are more structured close to the image manifold and are rather uniform farther away from the manifold.

% \paragraph{Two shuffled images}

% \paragraph{Decision surfaces across different optimizers.}

\section{Additional Reproducibility results}
\label{appendix_sec:repro}

\begin{figure*}[t]
\begin{center}
\includegraphics[width=.7\linewidth]{fig/dd/db_plots/legend_class.png}
\includegraphics[width=.8\linewidth]{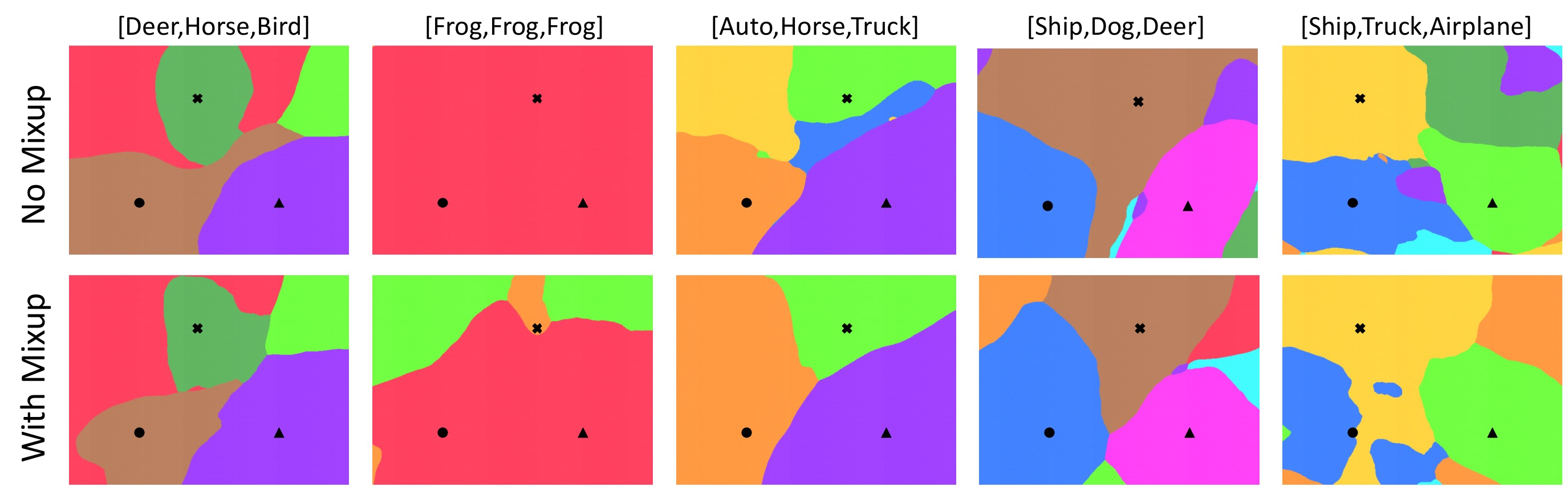}
\end{center}
\caption{
We present decision regions for random triplets sampled from the training set for ResNet18. We see the decision regions are almost same with and without mixup.}
\label{appendix_fig:db_withmixup}
\end{figure*}

\paragraph{With and without Mixup in training}
In order to understand how having mixup in the training affects the decision boundaries, we examined 2 cases, ResNet18 and Vision Transformer. In Fig.~\ref{appendix_fig:db_withmixup}, we show 5 randomly sampled triplets and their decision regions produced by ResNet18 trained with and without mixup. We can see there is a slight difference, but not quite significant. We quantified how ``similar" the decision surfaces are with reproducibility score introduced in Section 3.2. The score for Resnet18 is 0.774, and for ViT is 0.808. 

% \paragraph{Fragmentation scores across architectures}
% \begin{figure*}[h]
% \begin{center}
% \includegraphics[width=.99\linewidth]{appendix_fig/fragmentation_scores_across_archs.pdf}
% \end{center}
% \caption{Fragmentation scores across architectures \gs{to beautify}}
% \label{appendix_fig:fragmentation_scores_dbs}
% \end{figure*}
% \paragraph{Fragmentation scores across optimizers}
\section{Additional Double Descent results}
\label{appendix_sec:double_descent_results}
\paragraph{Additional error plots}

\begin{figure}
     \centering
     \begin{subfigure}[b]{\linewidth}
         \centering
         \includegraphics[width=\linewidth]{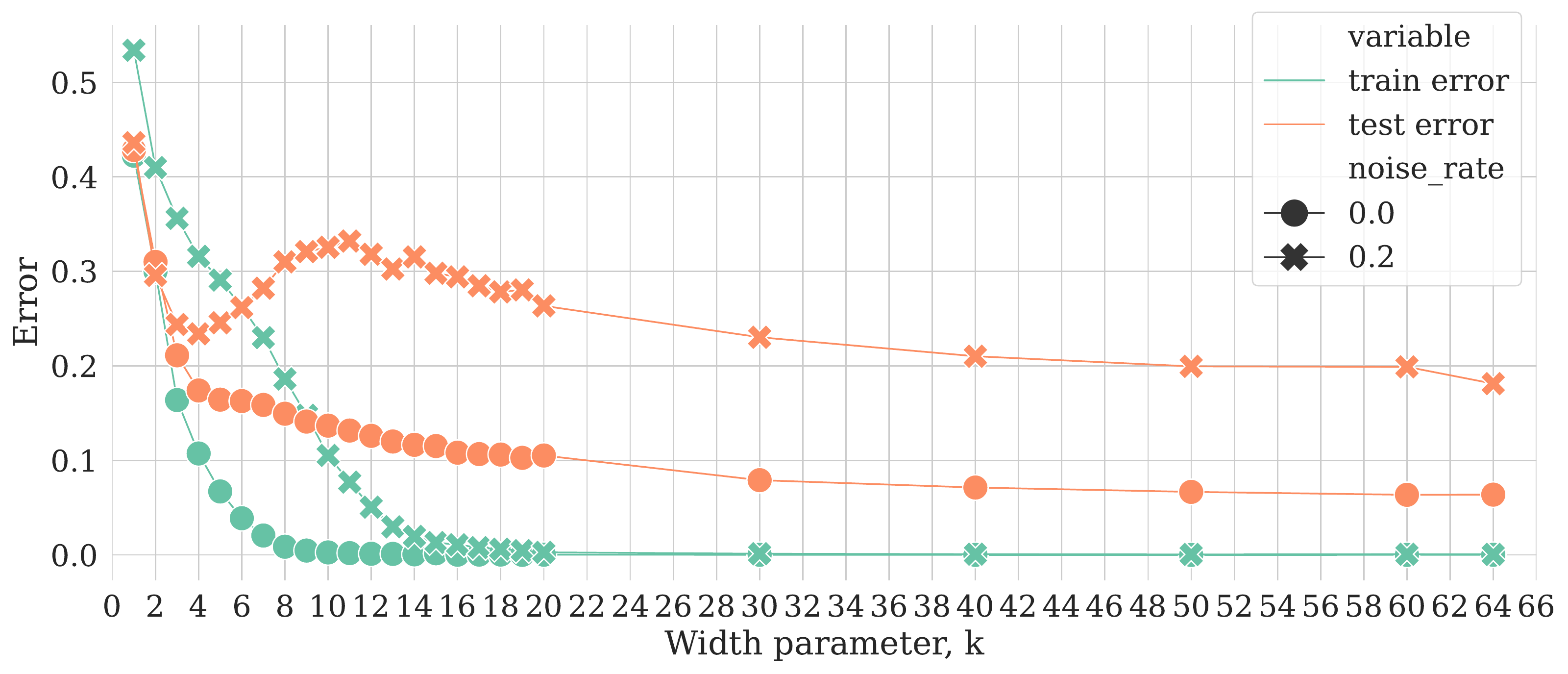}
     \end{subfigure}
    \vspace{-6mm}
\caption{In this figure we show the train and test errors with and without label noise.}
\label{appendix_fig:train_test_errors}
\end{figure}

\begin{figure}
     \centering
     \begin{subfigure}[b]{\linewidth}
         \centering
         \includegraphics[width=\linewidth]{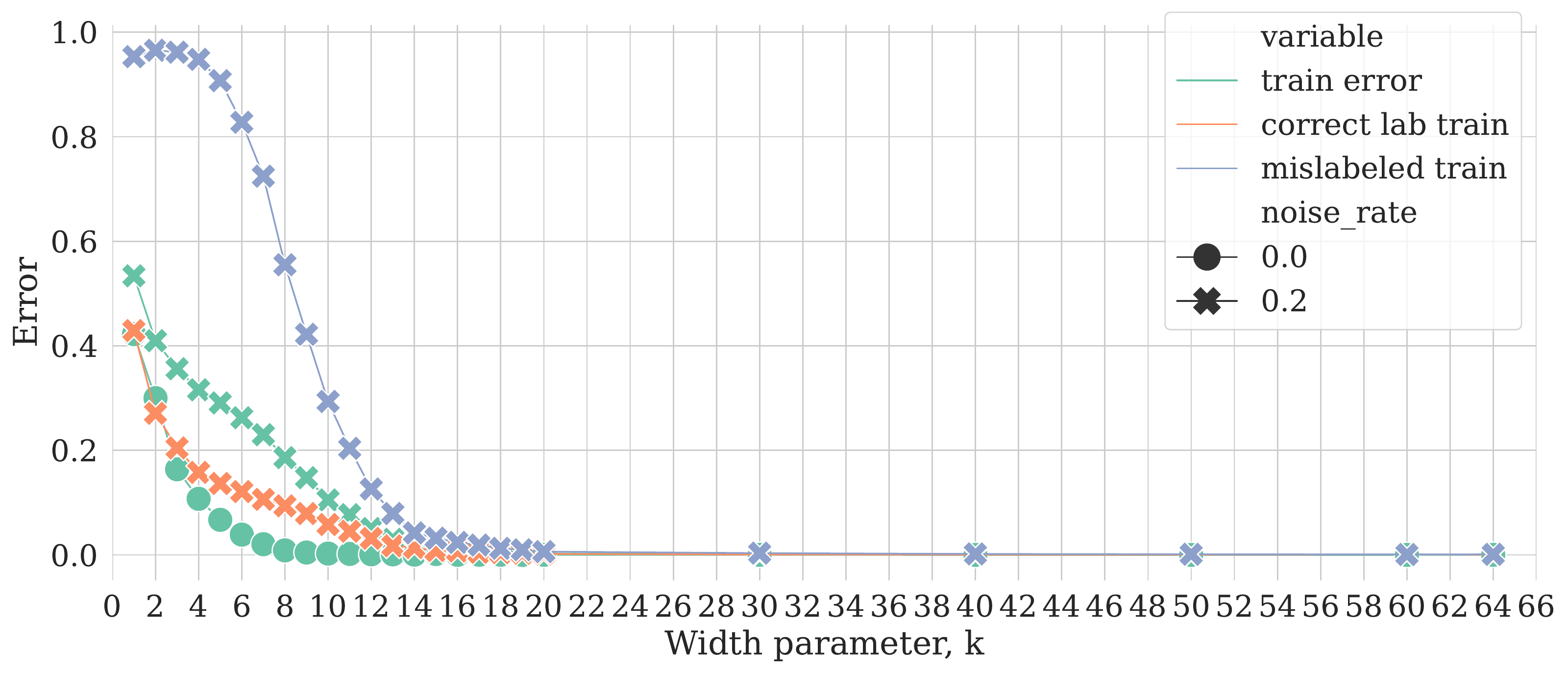}
     \end{subfigure}
    \vspace{-6mm}
\caption{In this figure, we show the train data errors for with and without label noise cases in green color. We also investigate how the errors are changing for correctly labeled points (orange curve) and in mislabeled points (grey curve). }
\label{appendix_fig:train_errors_breakdown}
\end{figure}

In Fig.~\ref{fig:dd_test_error}, we have seen how the test errors change as we progressively increase the model capacity. Figure~\ref{appendix_fig:train_test_errors} shows how training errors change in addition to test errors. We can see that the train error reaches 0 at much higher $k$ with label noise than without. In model without label noise, the interpolation begins at $k=10$ which is the true interpolation threshold when there is no label noise. We further examine how correctly labeled and mislabeled points are behaving in Figure~\ref{appendix_fig:train_errors_breakdown}. The green lines represent the overall train error, while orange shows the error on correctly labeled points. The mislabeled points are shown in grey, and the error is computed as incorrect predictions with respect to assigned class. We see that till $k=4$ the mislabeled points are not fit to their assigned class which partially explains the low test error of test data. However at $k=10$ most of the correctly labeled points are fit while some of the mislabeled points are still not fit to their assigned class. This trend diverges from what is seen in simple model families where the second peak of test error coincides with the model capacity with $0$ training error. This shows that double descent in more complicated in neural-network architectures than what is seen in simple linear models. 

\begin{figure}
     \centering
     \begin{subfigure}[b]{\linewidth}
         \centering
         \includegraphics[width=\linewidth]{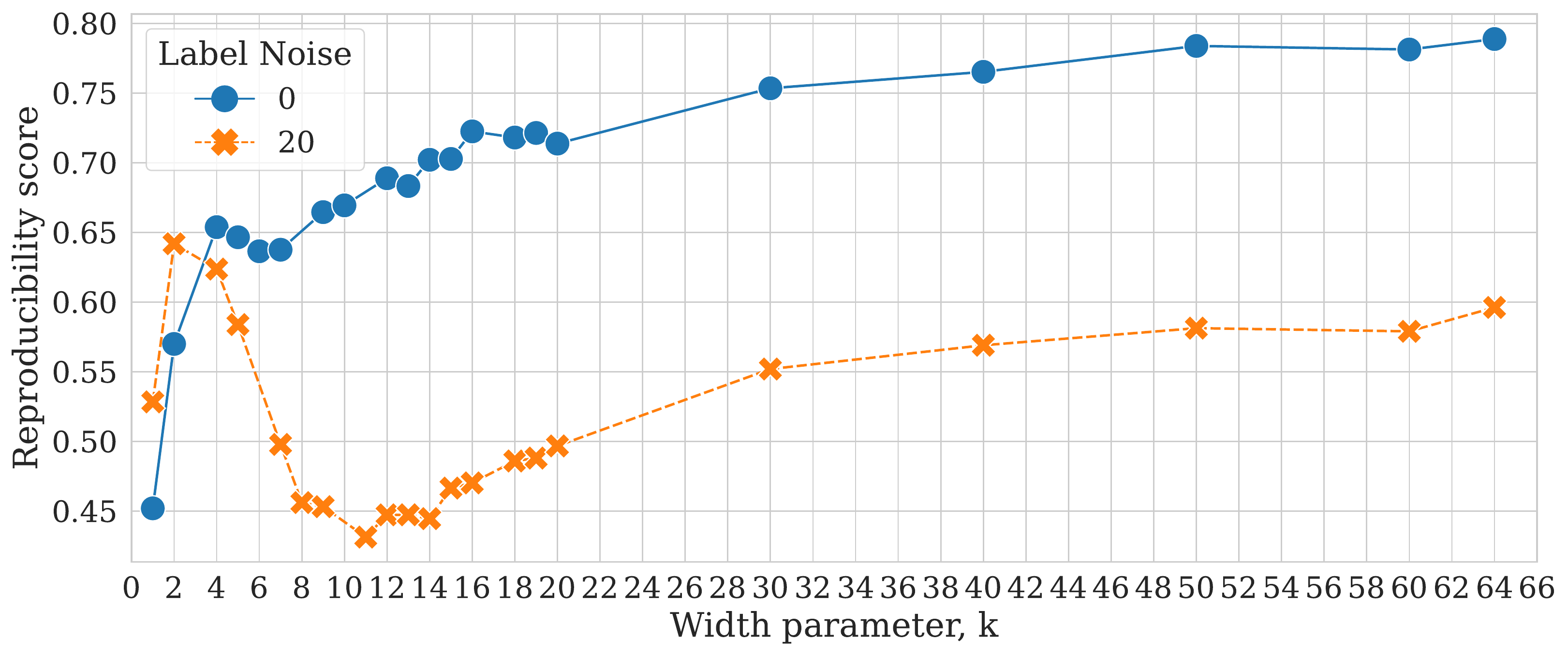}
     \end{subfigure}
    \vspace{-6mm}
\caption{Reproducibility with respect to random data samplings for models of different widths}
\label{appendix_fig:dd_repro_data}
\end{figure}

\vspace{-3mm}
\paragraph{Reproducibility scores from random data sampling}
In Figure~\ref{fig:dd_repro_all}, we have seen how decision boundaries change when we compare two runs of the same model architecture with different initializations. In Figure~\ref{appendix_fig:dd_repro_data}, we show how the ordering of the data changes the decision boundaries. We  see that the reproducibility across training runs is high in the under- and over-parametrized regimes, but it drops drastically closer to the interpolation threshold. This is the exact same behaviour observed in Figure~\ref{fig:dd_repro_all}. This shows that $k=10$ is a quite unstable with respect to different types of variations in the model training.

% \section{Decision boundaries wrt OOD samples}

\begin{figure*}
     \centering
     \begin{subfigure}[b]{.9\linewidth}
         \centering
         \includegraphics[width=.8\linewidth]{fig/dd/db_plots/legend_class.png}
\includegraphics[width=0.9\linewidth]{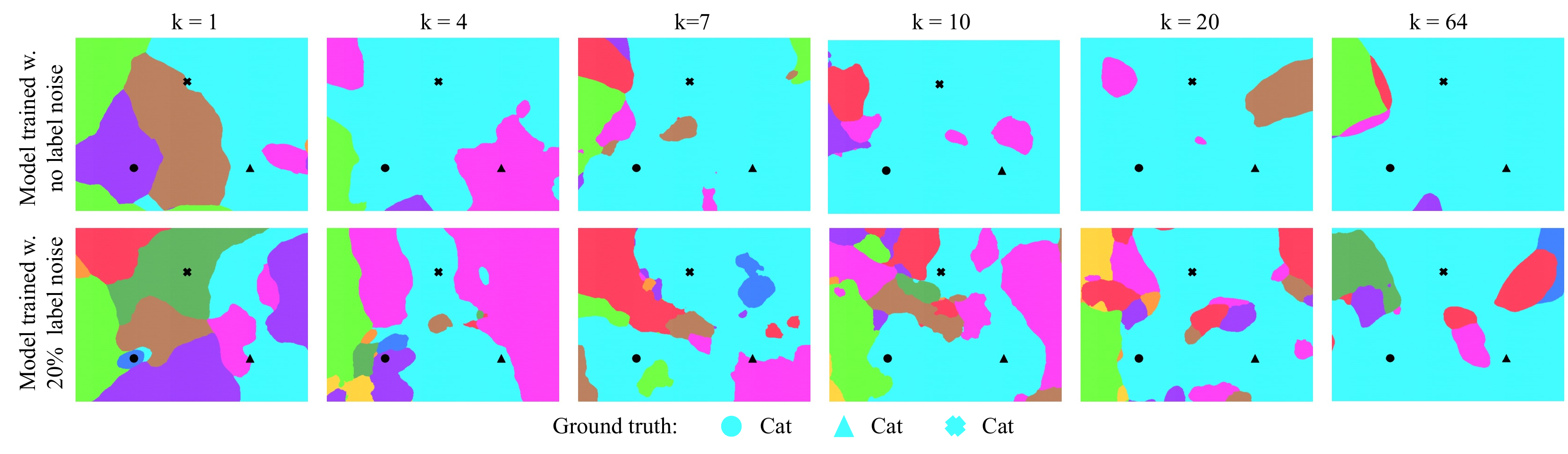}

         \caption{All points are from same class (Cat), and are correctly labeled even in label noise case.}
        %  \label{fig:dd_3random}
     \end{subfigure}
     \vfill
     \begin{subfigure}[b]{.9\linewidth}
         \centering
\includegraphics[width=0.9\linewidth]{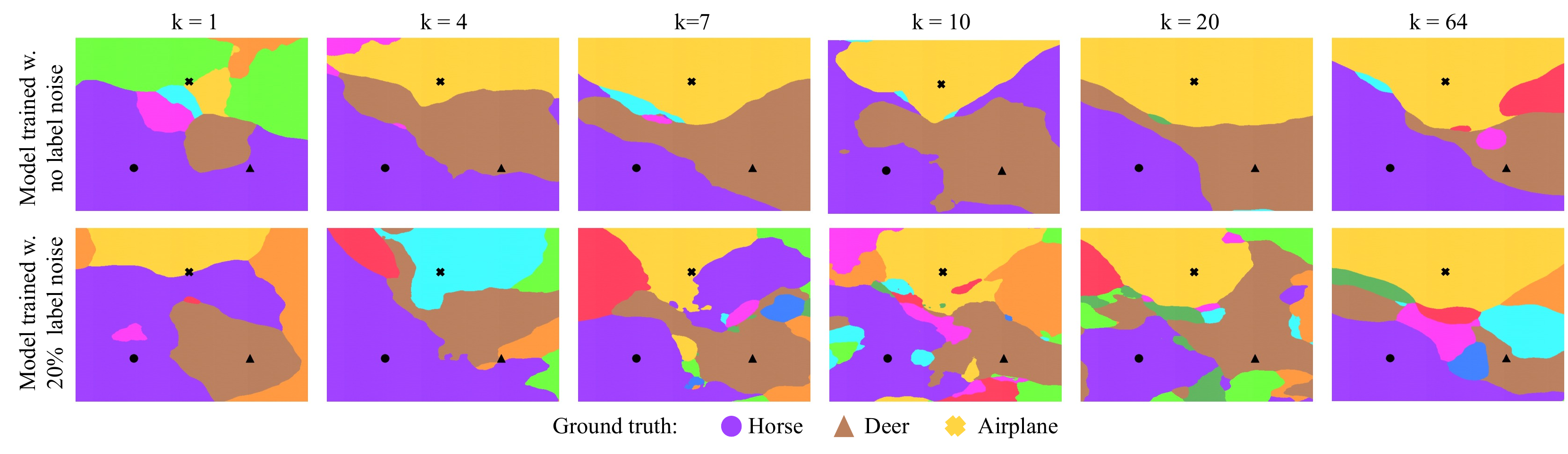}
         \caption{The images are sampled from 3 different classes and are correctly labeled.}
        %  \label{fig:dd_3corr}
     \end{subfigure}
     \vfill
     \begin{subfigure}[b]{.9\linewidth}
         \centering
\includegraphics[width=0.9\linewidth]{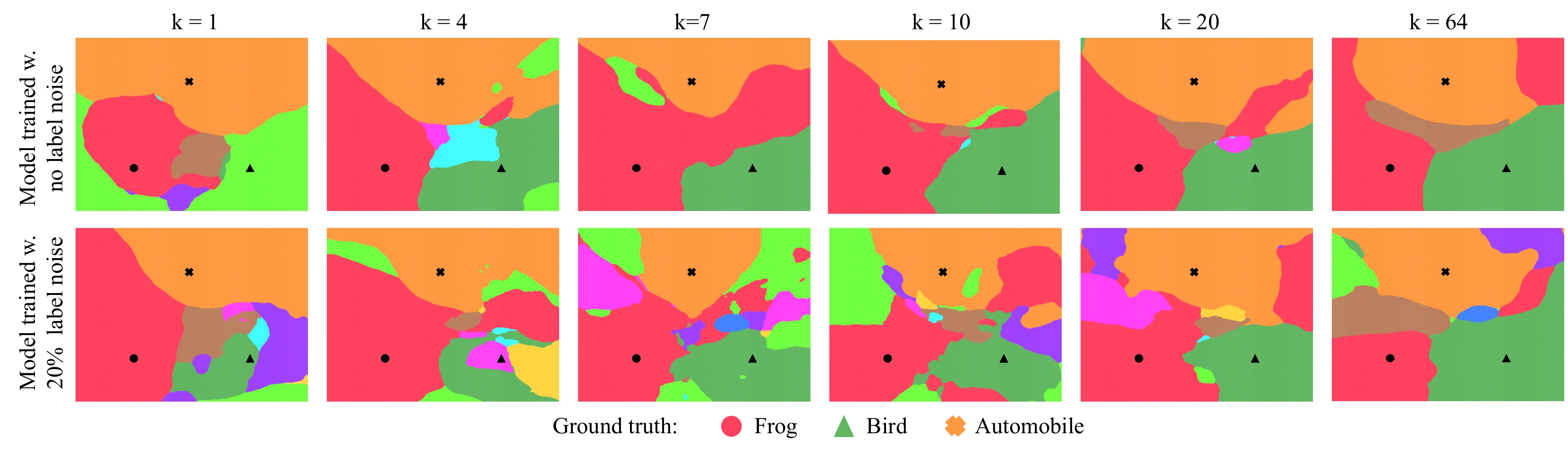}
         \caption{The images are sampled from 3 different classes and are correctly labeled. Additional case}
        %  \label{fig:dd_3corr}
     \end{subfigure}
     \vfill
     \begin{subfigure}[b]{.9\linewidth}
         \centering
\includegraphics[width=0.9\linewidth]{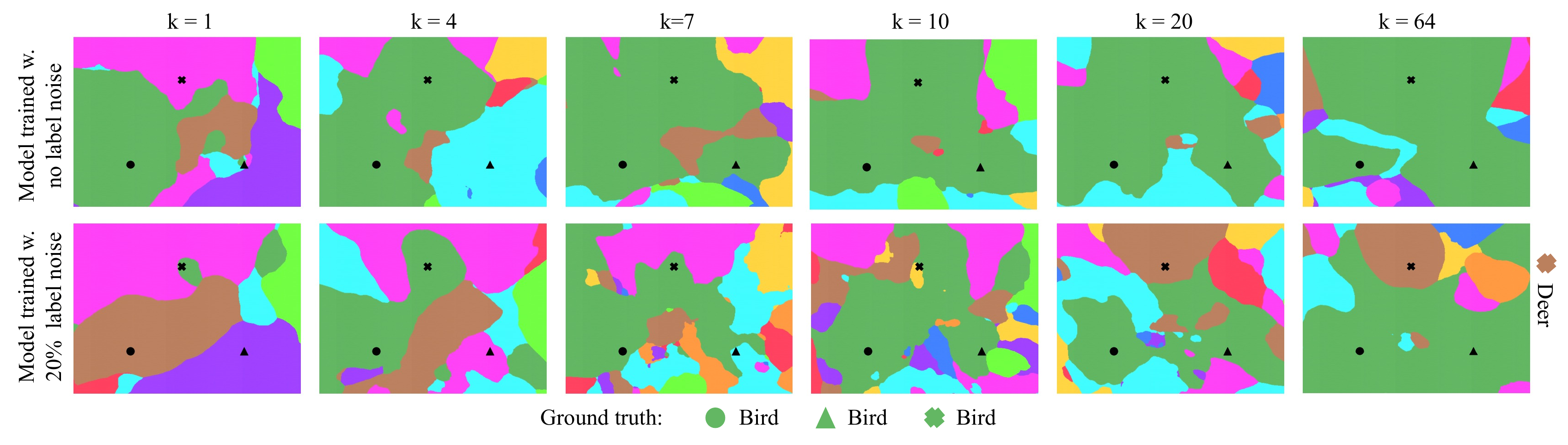}
         \caption{In this triplet, when there is no label noise, all three belonged to Bird class. But in the label noise case, the third point is mislabeled as Deer.}
        %  \label{fig:dd_3corr}
     \end{subfigure}
     \vfill
     \begin{subfigure}[b]{.9\linewidth}
         \centering
\includegraphics[width=0.9\linewidth]{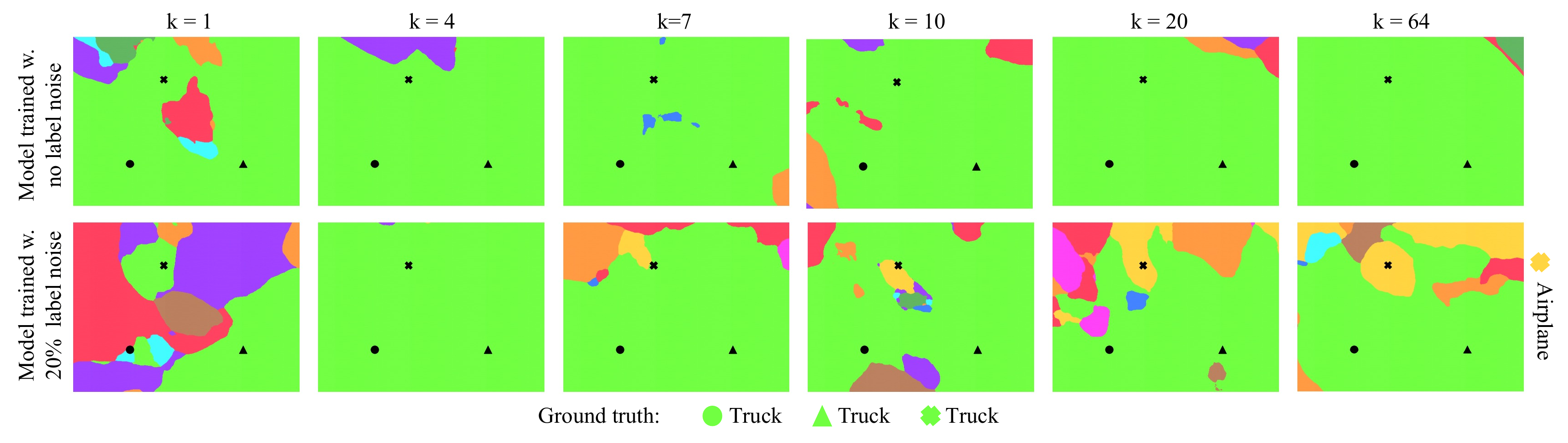}
         \caption{In this triplet, when there is no label noise, all three belonged to Truck class. But in the label noise case, the third point is mislabeled as Airplane.}
        %  \label{fig:dd_3corr}
     \end{subfigure}
\caption{\textbf{Decision boundaries for models of varying width.} We show additional decision surfaces with different types of triplets here.
\vspace{-4mm}
}
\label{appendix_fig:dd_additional_dbs}

\end{figure*}

\paragraph{Additional plots across varying model capacities and noise}
In Figure~\ref{appendix_fig:dd_additional_dbs}, we show how the decision regions change with and without label noise and with varying model capacities across different samplings of triplets.

\end{document}